%% file: main.tex
\documentclass[10pt,twocolumn,letterpaper]{article}

\usepackage{cvpr}      

\input{preamble}
\usepackage{caption}

\definecolor{cvprblue}{rgb}{0.21,0.49,0.74}
\definecolor{cityblue}{rgb}{0.21,0.49,0.74}
\usepackage[pagebackref,breaklinks,colorlinks,allcolors=cvprblue]{hyperref}

\def\paperID{*****} 
\def\confName{CVPR}
\def\confYear{2026}

\title{Aphanta: Diagnosing Task-Aligned Image-Edited Intermediates \\ for Multimodal Reasoning}

\author{
    Hengyuan Xu$^{1,2}$
    \quad
    Wei Cheng$^{2,\dag}$
    \quad
    Yumeng Ji$^{3}$
    \quad
    Xuanyang Zhang$^{2}$\\
    \quad
    Xianfang Zeng$^{2}$
    \quad
    Gang Yu$^{2,\ddag}$ 
    \quad
    Xingjun Ma$^{1,\ddag}$\\ [0.4em]
    $^{1}$ {Fudan University}
    \quad
    $^{2}$ {StepFun}
    \quad
    $^{3}$ {Shanghai Jiaotong Univeristy}
}

\begin{document}

\input{figures/teaser}
{\let\thefootnote\relax\footnotetext{\noindent$\dag$ Project lead; \ddag Corresponding authors.}}

\begin{abstract}
Explicit visual intermediates can help multimodal large language models (MLLMs) externalize spatial evidence and updated visual states, but their utility depends on whether an image editor can faithfully realize the required transformation. We introduce \textbf{Aphanta}, an automated task-discovery and closed-loop diagnostic framework for the $\text{MLLM}\rightarrow\text{image editor}\rightarrow\text{MLLM}$ pipeline. Aphanta evaluates three conditions---direct reasoning, reasoning with an editor-generated intermediate, and reasoning with an idealized reference intermediate---to separate potential visual headroom from the practical utility of current editors. Across 20 candidate tasks and multiple editor--MLLM combinations, we find that utility is strongly task-conditioned. Gains concentrate in visual cue injection, grounding, and counterfactual state realization, whereas intermediates requiring symbol-sensitive construction or structural extrapolation are substantially less reliable. On the selected positive-task subset, our consolidated Qwen pipeline improves the mean task score from 0.343 to 0.445 ($+10.2$ points; $+29.7\%$ relative), while the full study also retains filtered and unsuccessful tasks to expose the boundary. These results position image editing as a specialized visual workspace rather than a universal reasoning mechanism, and establish Aphanta as a reusable protocol for measuring task--representation alignment, editor realization, and downstream pipeline utility.
\end{abstract}

\input{sections/introduction}

\input{sections/related.tex}
\input{sections/method}

\input{sections/experiment}
\input{sections/conclusion}

\bibliographystyle{ieeenat_fullname}
\bibliography{main}

\clearpage
\appendix
\input{sections/appendix}

\end{document}

%% file: preamble.tex
\usepackage[utf8]{inputenc}
\usepackage[T1]{fontenc}
\usepackage{microtype}
\usepackage{amsmath}
\usepackage{amssymb}
\usepackage{amsfonts}
\usepackage{booktabs}
\usepackage{xspace}
\usepackage{enumitem}
\usepackage{multirow}
\usepackage{graphicx}
\usepackage{subcaption}
\usepackage{tabularx}
\usepackage{svg}
\usepackage{wrapfig}
\usepackage{fontawesome5}
\usepackage{colortbl}
\usepackage{tikz}
\usepackage{ulem}
\usepackage[most]{tcolorbox}
\usepackage{fancyvrb}
\usepackage{fvextra}
\usepackage{array}
\usepackage{url}
\usepackage{nicefrac}

\definecolor{boxpink}{HTML}{ECF8F4}
\definecolor{boxblue}{HTML}{EEF4FB}
\definecolor{boxorange}{HTML}{FFF4EA}
\definecolor{framepink}{HTML}{3E8E7E}
\definecolor{frameblue}{HTML}{4A6FA5}
\definecolor{frameorange}{HTML}{C97B3B}

\definecolor{tableHeader}{HTML}{264E63}
\definecolor{tableHeaderAlt}{HTML}{356B82}
\definecolor{tableSubheader}{HTML}{E8F0F4}
\definecolor{tableStripe}{HTML}{F4F7F9}
\definecolor{tableHighlight}{HTML}{E5F2EE}
\definecolor{tablePositive}{HTML}{167052}
\definecolor{tableNegative}{HTML}{B44949}
\definecolor{tableMuted}{HTML}{687983}

\newcommand{\tablehead}[1]{\textcolor{black}{\textbf{#1}}}
\newcommand{\tablesubhead}[1]{\textcolor{tableHeader}{\textbf{#1}}}
\newcommand{\posresult}[1]{\textcolor{tablePositive}{\bfseries #1}}
\newcommand{\negresult}[1]{\textcolor{tableNegative}{\bfseries #1}}

\newcommand{\opground}{\textcolor{colorGrounding}{\textbf{Grounding}}}
\newcommand{\opstructured}{\textcolor{colorLogic}{\textbf{Structured}}}
\newcommand{\opcue}{\textcolor{colorLowLevel}{\textbf{Cue}}}
\newcommand{\opstate}{\textcolor{colorEditing}{\textbf{State}}}

\definecolor{colorLowLevel}{HTML}{5575AD}
\definecolor{colorGrounding}{HTML}{3E8E7E}
\definecolor{colorEditing}{HTML}{C97B3B}
\definecolor{colorLogic}{HTML}{7B6D9C}

\newtcbox{\highlighttext}[2]{%
  on line,
  arc=4pt,
  colback=#2!80,
  boxrule=0pt,
  left=0.3pt, right=0pt, top=-1.5pt, bottom=-1.5pt,
  nobeforeafter,
  tcbox raise base,
  enhanced,
  fontupper=\color{#1}%
}

\newtcbox{\passbadge}{on line, arc=2pt, boxrule=0pt, colback=tableHighlight,
  left=1.6pt, right=1.6pt, top=0.4pt, bottom=0.4pt, nobeforeafter,
  tcbox raise base, fontupper=\scriptsize\bfseries\color{tablePositive}}
\newtcbox{\failbadge}{on line, arc=2pt, boxrule=0pt, colback=red!8,
  left=1.6pt, right=1.6pt, top=0.4pt, bottom=0.4pt, nobeforeafter,
  tcbox raise base, fontupper=\scriptsize\bfseries\color{tableNegative}}
\newcommand{\statuspass}{\passbadge{PASS}}
\newcommand{\statusfail}{\failbadge{FAIL}}

%% file: figures/teaser.tex

\twocolumn[{
  \renewcommand\twocolumn[1][]{#1}%
  \maketitle
  \begin{center}
    \vspace{-2.5ex}
    \includegraphics[width=0.98\textwidth,height=0.42\textheight,keepaspectratio]{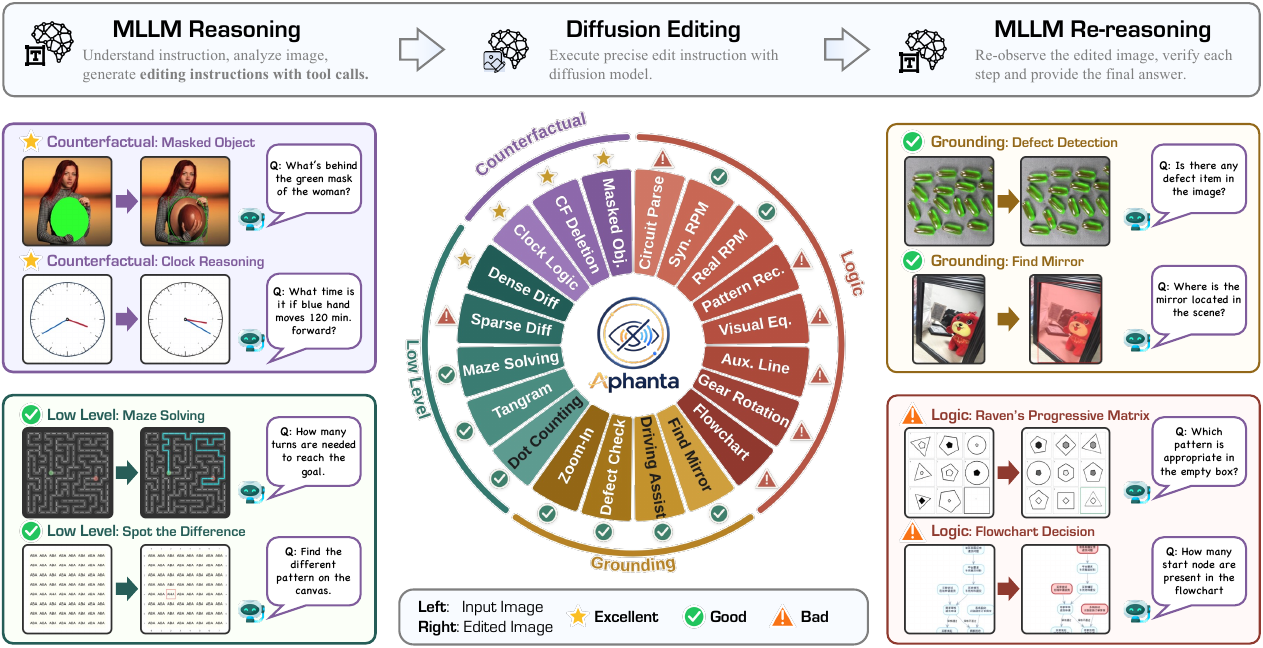}
    \vspace{-1.0ex}
    \captionsetup{hypcap=false}
    \captionof{figure}{\textbf{Task-Conditioned Utility of Image-Edited Intermediates.} Current instruction editors provide reliable assistance for selected cue-injection, grounding, and counterfactual state-realization tasks, but are less reliable when the requested intermediate requires exact structural extrapolation. Aphanta measures this boundary rather than assuming that one visual representation helps every task.}
    \label{fig:teaser}
    \vspace{-1.0ex}
  \end{center}
}]

%% file: sections/introduction.tex
\section{Introduction}
\label{sec:introduction}

The idea that a reasoning system can benefit from an auxiliary visual state predates current multimodal large language models (MLLMs). Earlier work used synthetic scenes, modular visual programs, and imagined rollouts to expose compositional structure or predict task-relevant future observations \cite{andreas2016neural,johnson2017clevr,johnson2017inferring,weber2017imagination,ebert2018visualforesight,ha2018worldmodels}. Modern MLLMs make this idea operational at inference time: they crop, zoom, mark, sketch, or otherwise transform an image to acquire evidence that was difficult to use in the original view \cite{zheng2025deepeyes,hong2025deepeyesv2,zhang2025thyme,lai2025mini-o3,qiao2025v}. A broader line of work generates explicit images, videos, or latent visual states as intermediate thoughts \cite{chern2025thinkingwithgeneratedimages,gu2025thinkmorph,he2025diffthinker,dai2026endocot,cheng2026omnir1unifiedgenerativeparadigm,yang2025machine,li2026thinking}. Together, these developments suggest a useful abstraction: an MLLM may benefit from a \emph{visual workspace} that makes a task-relevant state easier to perceive or reason over.

The existence of such a workspace, however, does not imply that every task benefits from an RGB intermediate. Image-to-image translation and instruction editing have become substantially more controllable \cite{isola2017image,zhu2017unpaired,zhu2016generative,meng2022sdedit,hertz2023prompt,brooks2023instructpix2pix,zhang2024magicbrush,liu2025step1x-edit}, but a visually plausible edit may still fail the specific count, relation, symbol, or counterfactual state required by a reasoning task. Three questions must therefore be distinguished. First, does an ideal intermediate visual state provide any \emph{headroom} over direct reasoning? Second, can a current image editor faithfully realize that state from an MLLM instruction? Third, does the downstream MLLM use the returned evidence rather than merely benefiting from an additional reasoning call or the text scaffold surrounding a tool invocation? Recent studies show that visual tool use is highly sensitive to task, representation, reliability, and cost \cite{yu2026avic,zhang2026adaptmmbench,li2026reliable,wang2026illusion,shao2026textcall}. Yet the practical utility of a standardized $\text{MLLM}\rightarrow\text{image editor}\rightarrow\text{MLLM}$ pipeline remains under-characterized across heterogeneous visual transformations.

We study this problem with \textbf{Aphanta}, an automated framework for task discovery, data construction, and closed-loop utility diagnosis. For each candidate task, Aphanta compares direct reasoning (A), reasoning with an editor-generated intermediate (B), and reasoning with an idealized reference intermediate (C). The comparison estimates whether a task admits useful visual assistance and whether the tested editor realizes it well enough to improve the complete pipeline. Repeating this procedure across 20 candidate tasks produces an empirical \emph{affordance map}: current instruction-conditioned editors are most reliable when injecting local cues, grounding relevant content, or realizing an explicitly requested visual state, but are less reliable when the intermediate requires symbol-sensitive construction or abstract structural extrapolation.

This framing changes the goal from deciding whether diffusion models ``can reason'' to measuring the alignment among a task, an intermediate representation, and an editor. It also avoids extrapolating from one editor to the entire diffusion or generative modeling paradigm. Specialized systems can learn visual planning and structured transformations under targeted supervision \cite{dai2026endocot,zhou2026visualplanning,sugiyama2026wisrd}; our results instead characterize current general instruction editors under the protocol and task suite studied here.

Our contributions are threefold:
\begin{itemize}[leftmargin=*,nosep]
    \item We introduce Aphanta, a reusable task-discovery and validation framework for measuring visual headroom and practical utility in an MLLM--image-editor--MLLM loop.
    \item We construct Aphanta Train and Test and report a 20-task diagnostic study spanning cue injection, grounding, counterfactual state realization, and structured visual construction, including filtered and unsuccessful tasks rather than only successful demonstrations.
    \item Across multiple MLLM--editor combinations and two external benchmarks, we identify a task-conditioned reliability boundary and translate it into concrete guidance for selective, verifiable visual assistance, while explicitly delimiting what the present A/B/C protocol can and cannot establish causally.
\end{itemize}

%% file: sections/related.tex
\section{Related Work}
\label{sec:related}

\subsection{Active and Generative Visual Reasoning}
Early visual-reasoning systems already treated intermediate visual structure as useful evidence. Neural module networks and program-execution models made the reasoning process explicit over synthetic or compositional scenes \cite{andreas2016neural,johnson2017clevr,johnson2017inferring}, while model-based agents used imagined rollouts or predicted future frames to support decisions \cite{weber2017imagination,ebert2018visualforesight,ha2018worldmodels}. Contemporary MLLMs revisit the same theme with stronger foundation models: active-perception methods crop, zoom, mark, or sketch the input during inference \cite{zheng2025deepeyes,hong2025deepeyesv2,zhang2025thyme,lai2025mini-o3,duan2025codeplot,qiao2025v}. A complementary family generates new visual states. Thinking with Generated Images, ThinkMorph, Omni-R1, DiffThinker, and EndoCoT interleave generation with reasoning or train generation models for structured visual tasks \cite{chern2025thinkingwithgeneratedimages,gu2025thinkmorph,cheng2026omnir1unifiedgenerativeparadigm,he2025diffthinker,dai2026endocot}. Other work uses video frames, continuous visual tokens, object blueprints, vector graphics, or language-native perception programs rather than edited RGB images \cite{li2026thinking,yang2025machine,li2025lvr,wang2025monetreasoninglatentvisual,ma2026thinking,yang2025omnisvg,janjua2026perceptionprograms}. These results establish that visual thought can take several forms; Aphanta asks when one particular form---an explicit image produced by a general instruction editor---has positive downstream utility.

\subsection{Selection, Reliability, and Evidence Use}
Recent work increasingly treats visual assistance as a conditional resource. AVIC controls when and how much to imagine, AdaptMMBench separates mode-selection quality from final accuracy, and Gen-VCoT routes among RGB intermediates of different depth \cite{yu2026avic,zhang2026adaptmmbench,zhou2026genvcot}. ToolVision similarly aligns supervision with the learner's own tool benefit \cite{mao2026toolvision}. Reliability is equally important: Reliable Thinking with Images filters noisy cues, while process-reward and evidence-grounding benchmarks diagnose errors that outcome accuracy alone cannot reveal \cite{li2026reliable,zhou2026prm,li2026viebench}. MentisOculi further shows that even reference visualizations need not improve reasoning \cite{zeller2026mentisoculi}. Causal audits show that a model may call a tool without using its returned evidence, and that structured text emitted before a tool call can sometimes account for much of the observed gain \cite{wang2026illusion,xiang2026openvistool,shao2026textcall}. Aphanta is complementary to these policy- and trajectory-level methods: it provides an editor-specific, task-level utility map, but does not treat an accuracy change alone as proof that the returned pixels were causally decisive.

\subsection{Benchmarks and Image Editors}
MIRA, BabyVision, VisuLogic, VisualPuzzles, and VSP expose persistent gaps in perception, spatial reasoning, and the use of auxiliary visual states \cite{zhou2025visualizingstepreasoningmira,chen2026babyvisionvisualreasoninglanguage,xu2025visulogic,song2025visualpuzzlesdecouplingmultimodalreasoning,wu2025vsp}. ViEBench goes beyond final answers by checking whether a model grounds and uses the correct evidence \cite{li2026viebench}. Image editing, meanwhile, has progressed from paired and unpaired image-to-image translation to diffusion-based and instruction-guided editing \cite{isola2017image,zhu2017unpaired,meng2022sdedit,hertz2023prompt,brooks2023instructpix2pix,zhang2024magicbrush}. Recent editor-centric work studies practical instruction editing, reasoning-aware editing, region-adaptive generation, restoration through large editing models, and human-aligned evaluation \cite{liu2025step1x-edit,yin2025reasonedit,chen2025regione,yang2026realrestorer,wu2025kris,jiang2026geditbench}. Related generation benchmarks and systems further test nuanced image generation, identity consistency, many-to-many image manipulation, vector graphics, and story visualization \cite{chang2025oneigbench,xu2025withanyone,fu2025imontage,yang2025omnisvg,zhuang2026vistorybench}. Targeted studies also show that editor behavior is trainable: specialized models can learn visual planning or image-space rule execution \cite{zhou2026visualplanning,sugiyama2026wisrd}. Unlike an editor leaderboard or a benchmark that supplies only reference intermediates, Aphanta evaluates the closed loop from task proposal through actual editing to downstream reasoning, and contrasts practical edits with idealized reference states across a heterogeneous task pool.

%% file: sections/method.tex
\section{Aphanta: Diagnosing Visual-Intermediate Utility}
\label{sec:method}

\input{figures/pipeline}

We present \textbf{Aphanta}, an automated pipeline for discovering and validating tasks in which image-edited intermediates may change MLLM performance. Aphanta combines a standardized inference chain, a three-condition diagnostic protocol, and a four-phase task-development loop with human quality control.

\subsection{Research Setting}
\label{subsec:method_setting}

Our question is deliberately scoped to current instruction-conditioned image editors: for which tasks, and to what extent, does an editor-generated RGB intermediate improve a complete MLLM reasoning pipeline? We study the fixed chain
\[
q,x \rightarrow \mathrm{MLLM} \rightarrow e \rightarrow \mathrm{Editor}(x,e)
\rightarrow \tilde{x} \rightarrow \mathrm{MLLM} \rightarrow \hat{y},
\]
where $q$ is a query, $x$ the original image, $e$ an edit instruction, $\tilde{x}$ the edited intermediate, and $\hat{y}$ the final response. The editor may be called iteratively. Unless stated otherwise, the same MLLM produces $e$ and answers after editing, so model-family changes do not confound the two reasoning stages within a pipeline.

We treat the editor as a candidate \emph{visual workspace}, not as an autonomous solver. This distinction matters because a correct instruction can be rendered incorrectly, and a plausible rendering can still be unhelpful to the downstream MLLM.

\subsection{Three-Condition Utility Diagnosis}
\label{subsec:method_selection}

For each task, Aphanta evaluates three conditions under a shared scoring function:
\begin{itemize}[leftmargin=*,nosep]
    \item \textbf{A (Direct):} the MLLM answers from $(q,x)$ without an edited intermediate;
    \item \textbf{B (Actual Edit):} the full pipeline generates $e$ and $\tilde{x}$ before re-reasoning;
    \item \textbf{C (Reference):} the generated image is replaced with an idealized, programmatically constructed reference intermediate.
\end{itemize}

Let $S_A,S_B,S_C$ be task scores in the three conditions. We report
\[
\Delta_{\mathrm{edit}}=S_B-S_A, \qquad
\Delta_{\mathrm{ref}}=S_C-S_A.
\]
$\Delta_{\mathrm{ref}}$ is diagnostic visual headroom: it asks whether the downstream MLLM can benefit from the intended intermediate. $\Delta_{\mathrm{edit}}$ measures the practical utility of the complete editor-in-the-loop pipeline. A large positive $\Delta_{\mathrm{ref}}$ with a small or negative $\Delta_{\mathrm{edit}}$ identifies a realization gap under the tested pipeline. We call C a \emph{reference}, rather than a strict upper bound, because an actual edit may occasionally interact with the downstream model more favorably than the constructed reference.

This triad does not, by itself, prove that the rendered pixels causally determine the answer: B also contains the edit instruction and an additional MLLM turn. We therefore interpret $\Delta_{\mathrm{edit}}$ as pipeline-level utility and return to matched-call, no-return, and sham-image controls in \autoref{subsec:limitations}.

\subsection{Automated Task-Development Loop}
\label{subsec:method_pipeline}

To scale task discovery, we build a four-phase agent pipeline. The agent reuses prompt templates, data utilities, and training/evaluation scripts; human experts supervise feasibility and data quality.

\textbf{Phase 1: Proposal and screening.} Given the research question and previously observed task patterns, the agent proposes candidates that require a non-trivial visual transformation or disambiguation. Human reviewers screen conceptual feasibility, measurable evaluation, and the availability of constructible reference intermediates.

\textbf{Phase 2: Preliminary diagnosis.} The agent builds a small validation set from public data, web resources, or procedural synthesis and evaluates A/B/C. Tasks with negligible reference headroom, saturated baselines, unavailable edited outputs, or unstable formulations may be stopped; all such outcomes remain in the reported 20-task audit rather than disappearing from the study.

\textbf{Phase 3: Data construction and curation.} For retained tasks, the agent collects or synthesizes data, creates task-specific prompts and reference intermediates, and produces visual summaries. Human reviewers check relevance, diversity, and construction noise before training.

\textbf{Phase 4: Training, evaluation, and iteration.} The agent launches task-specific or consolidated editor training and evaluates the resulting pipeline. Results guide prompt, formatting, and training updates and are also fed back to Phase 1, closing the loop between hypothesis formation and quantitative validation.

\subsection{Task Pool and Outputs}
\label{subsec:method_outputs}

Running the loop yields \textbf{Aphanta Train}, a collection of task-specific editing resources, and \textbf{Aphanta Test}, a standardized evaluation pool for the same inference chain. We organize tasks by their dominant requested operation: grounding, perceptual cue injection, counterfactual state realization, or structured extrapolation. These labels summarize the dominant operation and are not claimed to be mutually exclusive cognitive categories. Crucially, the task pool preserves low-headroom, stopped, and unsuccessful cases, enabling a more realistic map than a success-only collection. Examples are shown in \autoref{fig:subtasks_data}.

\input{figures/subtasks_data}

%% file: figures/pipeline.tex
\begin{figure*}[htbp]
\vspace{-2em}
    \centering
    \includegraphics[width=\textwidth]{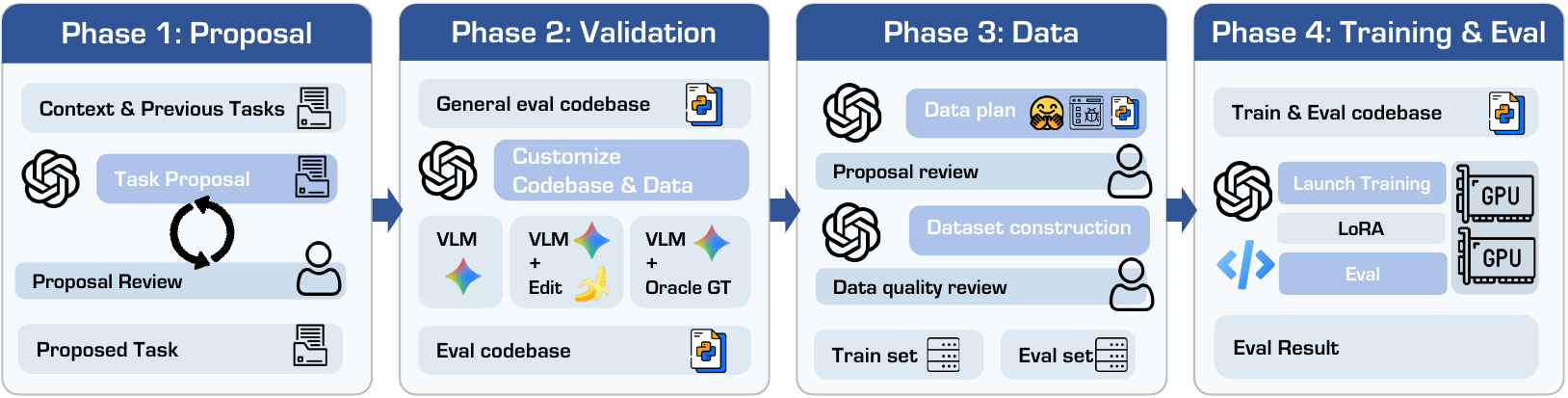}
    \caption{\textbf{The Aphanta Pipeline.} A four-phase loop proposes and screens tasks, performs preliminary A/B/C diagnosis, constructs and reviews data, and trains and evaluates editor-in-the-loop pipelines. Results feed back into task proposal, producing a task-conditioned utility map rather than a success-only collection.}
    
    \label{fig:pipeline}
    \vspace{-2em}
\end{figure*}

%% file: figures/subtasks_data.tex
\begin{figure*}[htbp]
    \centering
    \includegraphics[width=\textwidth]{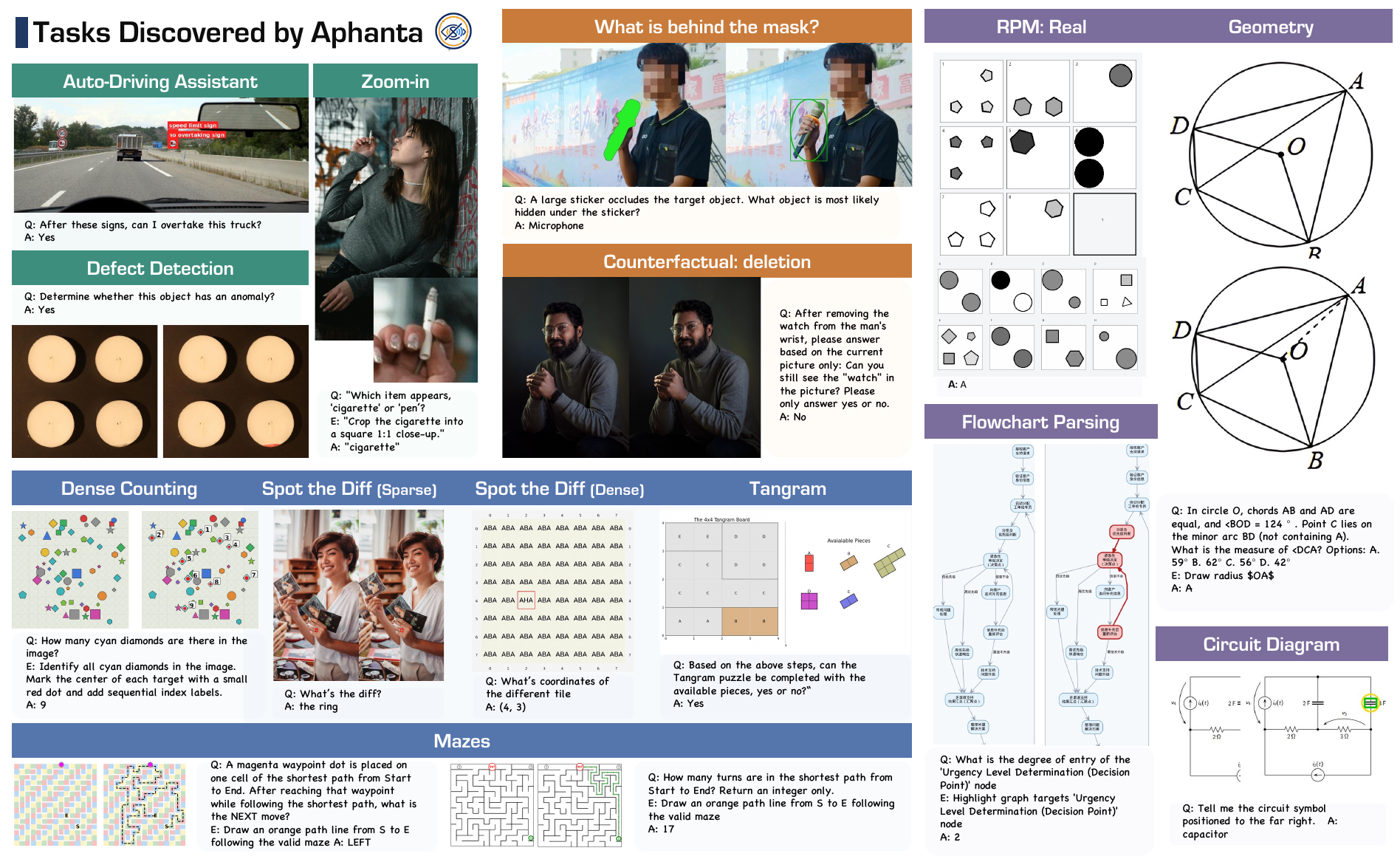}
    \caption{\textbf{Showcases of Tasks Aphanta Found.} The colors represent four different categories: \highlighttext{white}{colorLowLevel}{Low-level Recognition}, \highlighttext{white}{colorGrounding}{Grounding}, \highlighttext{white}{colorEditing}{Counterfactual}, and \highlighttext{white}{colorLogic}{Logic}.}
    \label{fig:subtasks_data}
\end{figure*}

%% file: sections/experiment.tex
\section{Task-Conditioned Utility of Edited Intermediates}
\label{sec:experiment}

\input{tables/subtasks}

\subsection{A 20-Task Affordance Map}

Table~\ref{tab:sub_tasks_summary} reports all 20 candidates considered by Aphanta, including stopped and unsuccessful tasks. Four tasks---\textit{Repeated Pattern Recognition}, \textit{Visual Equation Puzzle}, \textit{Gear Rotation Reasoning}, and \textit{Spot-the-Difference (Sparse)}---were stopped after preliminary diagnosis because of limited reference headroom, a saturated baseline, unavailable edited outputs, or insufficiently stable task construction. Of the 16 tasks taken forward, 15 reached final-stage evaluation; \textit{Circuit Diagram Parsing} retains its Phase-2 result because agent failed its development. Thirteen tasks yielded a positive, retained pipeline, whereas \textit{Plane Geometry Auxiliary Line} and \textit{Flowchart Decision} did not provide reliable final-stage assistance.

The task-level results reveal a graded boundary rather than a binary division between perception and logic. Counterfactual state realization is the clearest positive region: editing clocks, deleting specified objects, or completing a masked state increases the downstream score by 0.21--0.37. Cue-injection tasks such as dense counting, tangram decomposition, and dense difference marking also improve substantially. Grounding tasks generally benefit, although the actual edit can occasionally exceed the constructed reference, confirming that C is a diagnostic target rather than a strict numerical ceiling.

Tasks requiring exact structural extrapolation are less consistent. The two RPM variants improve after task-specific development, showing that a broad claim of ``no visual logic'' is not supported. In contrast, circuit symbols, geometry constructions, gear relations, and flowchart routing expose recurring realization errors. These tasks often retain positive reference headroom while the actual edit is weak or harmful, which is the signature of a task--editor mismatch under our protocol.

\input{tables/topics}

Table~\ref{tab:topic_average} summarizes the same pattern using unweighted macro-averages over tasks for which all A/B/C conditions are available. Because the task scores include both accuracy and IoU, the aggregation is descriptive rather than a universal metric or hypothesis test. State realization has the largest mean gain, followed by cue injection and grounding; the structured group has a negative mean actual-edit delta despite positive reference headroom. The group names denote dominant visual operations and should not be read as mutually exclusive cognitive categories.

\subsection{Consolidation and Cross-Model Transfer}

\input{tables/main_results_compare}
\input{figures/overall_compare}

We consolidate the retained training assets into a unified Qwen-Image-Edit model and evaluate the selected positive-task subset under a common protocol. As shown in Table~\ref{tab:main_results_compare}, the mean task score increases from 0.343 to 0.445: an absolute change of $+0.102$ ($+10.2$ percentage points on the normalized scale) and a $+29.7\%$ relative gain. This number describes the post-selection positive region; it is not an all-task editor leaderboard. The corresponding reference score of 0.558 indicates remaining realization headroom.

We next vary model families and editors. For within-family pipelines, Seed and Gemini obtain smaller positive deltas, whereas the GPT pairing is negative on this subset. Holding Qwen3-VL fixed and changing only the editor produces deltas from $-4.7$ to $+10.2$ points. This spread shows that downstream utility depends on editor execution, not only on the requested operation. It does not, however, fully isolate instruction quality, because proprietary systems and family-specific pipelines may differ in prompting, preprocessing, or hidden correction mechanisms.

The heatmap in \autoref{fig:overall_compare} further shows that no editor dominates every operation. Positive cells concentrate in state realization and selected grounding tasks; structured tasks and some seemingly simple cue operations remain unstable. Relative percentages are visually useful for comparing directions, but can be large when the direct baseline is small; the absolute deltas in Table~\ref{tab:main_results_compare} are therefore our primary summary.

\input{figures/goodcases}

\subsection{What Explains the Observed Boundary?}

The task study suggests three recurring properties of current instruction-conditioned editors. We state them as empirical properties of the tested systems, not as architectural impossibility results.

\begin{figure}[t]
    \begin{tcolorbox}[colback=white, colframe=black, arc=2mm, boxrule=1pt, top=2mm, bottom=2mm, left=2mm, right=2mm]
        \textbf{Finding 1: Reliable visual-state realization.}\\
        Editors are most useful when the requested intermediate can be expressed as a local addition, deletion, attribute change, or perceptual cue while preserving the remaining scene.
        \par\smallskip
        \centering
        \includegraphics[width=\linewidth]{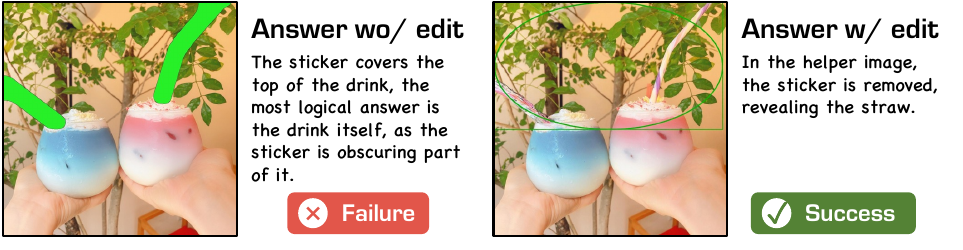}
    \end{tcolorbox}
    \caption{\textbf{Reliable Visual-State Realization.} A counterfactual task in which editing realizes the queried visual state before the final MLLM decision.}
    \label{fig:insight_1}
    \vspace{-1em}
\end{figure}

In counterfactual tasks, the original image can contain evidence that is inconsistent with the state described by the query. An editor can reduce this mismatch by rendering the requested state directly. The downstream model then answers from a representation aligned with the question instead of maintaining the update only in text. This mechanism is consistent with the strong gains for clock manipulation, deletion, and masked-state completion.

\begin{figure}[t]
    \begin{tcolorbox}[colback=white, colframe=black, arc=2mm, boxrule=1pt, top=2mm, bottom=2mm, left=2mm, right=2mm]
        \textbf{Finding 2: Plausibility can exceed task fidelity.}\\
        An intermediate may look coherent while violating a count, symbol, correspondence, or geometric constraint required by the task.
        \par\smallskip
        \centering
        \includegraphics[width=\linewidth]{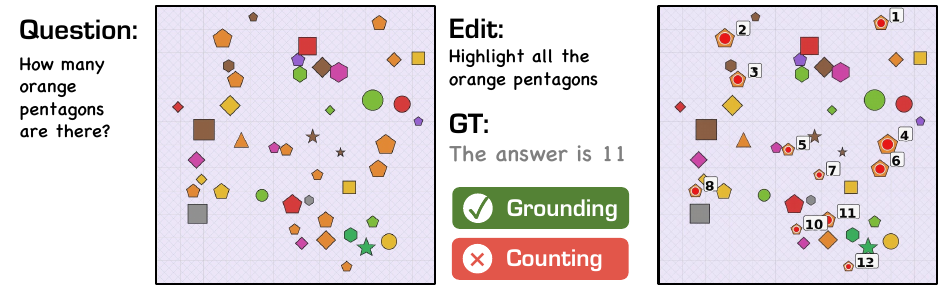}
    \end{tcolorbox}
    \caption{\textbf{Plausibility Can Exceed Task Fidelity.} A visually coherent numbering edit that omits a task-critical item.}
    \label{fig:insight_2}
    \vspace{-1em}
\end{figure}

This gap explains why an apparently polished intermediate can reduce final accuracy. In the example above, sequential labels are rendered naturally but one item is skipped. Once the downstream MLLM conditions on this image, the error can propagate. Similar error accumulation has been documented for noisy visual thoughts and mis-grounded visual-tool trajectories \cite{li2026reliable,wang2026illusion}.

\begin{figure}[t]
    \begin{tcolorbox}[colback=white, colframe=black, arc=2mm, boxrule=1pt, top=2mm, bottom=2mm, left=2mm, right=2mm]
        \textbf{Finding 3: Structured extrapolation is less reliable.}\\
        Current general editors struggle when success requires exact symbol recognition, logically valid construction, or topology-preserving routing beyond a locally specified edit.
        \par\smallskip
        \centering
        \includegraphics[width=\linewidth]{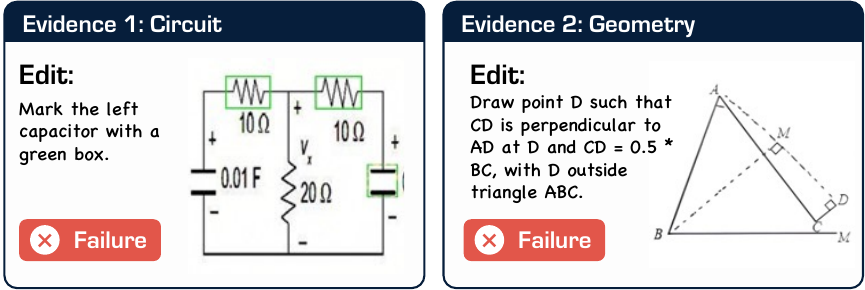}
    \end{tcolorbox}
    \caption{\textbf{Structured Extrapolation Is Less Reliable.} Representative realization errors in circuit interpretation and geometry construction.}
    \label{fig:insight_3}
    \vspace{-1em}
\end{figure}

The circuit, geometry, and flowchart cases require more than perceptual plausibility: every symbol, relation, or auxiliary construction must satisfy a discrete constraint. The tested instruction editors do not realize these transformations consistently. This is an empirical boundary of general editors under our setup, not a claim that diffusion or image-space models are fundamentally unable to learn such operations. Targeted training has already improved maze planning, Sudoku, and other structured transformations \cite{dai2026endocot,zhou2026visualplanning,sugiyama2026wisrd}.

\subsection{Scope, Causality, and Limitations}
\label{subsec:limitations}

Our conclusions are bounded in four ways. First, Aphanta is an exploratory task-discovery process; selection based on Phase-2 headroom can increase the magnitude of gains on the retained subset. We therefore report all 20 candidates and treat the consolidated positive-region score separately from the full task audit. Second, the operation labels overlap, and a larger preregistered task universe is needed before interpreting their macro-averages as population estimates. Third, closed-source editors expose limited architectural and decoding details, so cross-family results characterize products and APIs at evaluation time rather than isolated model components.

Finally, A/B/C establishes pipeline-level utility but not the causal contribution of returned pixels. Recent work finds that tool-call text can be sufficient in some visual-tool settings and that rendered visual thoughts may be attended to without contributing content \cite{li2026visualopsd,wang2026illusion,shao2026textcall}. A complete causal decomposition should add matched second-pass reasoning, call-without-return, identity or sham images, and step-level counterfactual replacements. These controls are a necessary next step for separating instruction scaffolding, editor realization, and downstream evidence use.

%% file: tables/subtasks.tex
\begin{table*}[t]
    \centering
    \caption{\textbf{Complete 20-Task Audit.} A/B/C denote Direct, Actual Edit, and Reference. \textit{Italicized scores} are preliminary values for tasks without final-stage evaluation; ``status'' records whether a task produced a retained practical pipeline, not whether visual assistance is possible in principle.}
    \label{tab:sub_tasks_summary}
    \scriptsize
    \setlength{\tabcolsep}{3pt}
    \renewcommand{\arraystretch}{1.06}
    \resizebox{\textwidth}{!}{%
    \begin{tabular}{c c c c c c c | c c c c c c c}
    \toprule
    \multirow{2}[2]{*}{\textbf{ID}} & \multirow{2}[2]{*}{\textbf{Task}} & \multirow{2}[2]{*}{\textbf{Status}} & \multirow{2}[2]{*}{\textbf{Topic}} & \multicolumn{3}{c|}{\textbf{Evaluation}} & \multirow{2}[2]{*}{\textbf{ID}} & \multirow{2}[2]{*}{\textbf{Task}} & \multirow{2}[2]{*}{\textbf{Status}} & \multirow{2}[2]{*}{\textbf{Topic}} & \multicolumn{3}{c}{\textbf{Evaluation}} \\
    \cmidrule(lr){5-7} \cmidrule(l){12-14}
    & & & & \textbf{Direct} & \textbf{+Edit} & \textbf{+Ref.} & & & & & \textbf{Direct} & \textbf{+Edit} & \textbf{+Ref.} \\
    \midrule
    1 & Where Is My Mirror & \statuspass & Grounding & 0.70 & 0.78 & 0.77 & 11 & Plane Geometry Aux. Line & \statusfail & Logic & 0.41 & 0.43 & 0.47 \\
    2 & Circuit Diagram Parsing & \statusfail & Logic & \textit{0.70} & \textit{0.60} & \textit{0.88} & 12 & Dense Dot Counting & \statuspass & Low Level & 0.05 & 0.23 & 0.52 \\
    3 & Auto. Driving Assistant & \statuspass & Grounding & 0.42 & 0.57 & 0.45 & 13 & Tangram & \statuspass & Low Level & 0.31 & 0.51 & 0.61 \\
    4 & RPM: Synthetic & \statuspass & Logic & 0.18 & 0.30 & 0.39 & 14 & Gear Rotation Reasoning & \statusfail & Logic & \textit{0.92} & \textit{0.42} & \textit{0.83} \\
    5 & RPM: Real & \statuspass & Logic & 0.61 & 0.79 & 0.71 & 15 & Industrial Defect Inspect & \statuspass & Grounding & 0.70 & 0.82 & 0.88 \\
    6 & Repeated Pattern Recog. & \statusfail & Logic & \textit{0.13} & \textit{0.07} & \textit{0.20} & 16 & Flowchart Decision & \statusfail & Logic & 0.36 & 0.22 & 0.41 \\
    7 & Analog Clock Reasoning & \statuspass & Counterfactual & 0.50 & 0.81 & 0.84 & 17 & Zoom-in (Remake) & \statuspass & Grounding & 0.87 & 0.92 & 0.93 \\
    8 & Counterfactual: Deletion & \statuspass & Counterfactual & 0.10 & 0.47 & 0.92 & 18 & Maze Solving & \statuspass & Low Level & 0.24 & 0.26 & 0.27 \\
    9 & What Is Behind the Mask? & \statuspass & Counterfactual & 0.25 & 0.46 & 0.68 & 19 & Spot-the-Diff (Sparse) & \statusfail & Low Level & \textit{0.53} & \textit{--} & \textit{0.58} \\
    10 & Visual Equation Puzzle & \statusfail & Logic & \textit{0.85} & \textit{--} & \textit{0.85} & 20 & Spot-the-Diff (Dense) & \statuspass & Low Level & 0.18 & 0.46 & 0.68 \\
    \bottomrule
    \end{tabular}%
    }
    \vspace{-2em}
\end{table*}

%% file: tables/topics.tex
\begin{table*}[htbp]
\vspace{-1em}
    \centering
    \begin{minipage}[c]{0.58\textwidth}
        \centering
        \captionof{table}{\textbf{Descriptive Macro-Average by Dominant Operation.} Each row uses tasks with all A/B/C scores available; stopped tasks use their preliminary scores.}
        \label{tab:topic_average}
        \scalebox{0.90}{
            \renewcommand{\arraystretch}{1.22}
            \setlength{\tabcolsep}{5pt}
            \begin{tabular}{@{}l|rrr|rr@{}}
            \toprule
            \multirow{2}{*}{\tablehead{Operation}} & \multicolumn{3}{c|}{\tablehead{Mean task score}} & \multicolumn{2}{c}{\tablehead{Absolute delta}} \\
             & \tablesubhead{Direct} & \tablesubhead{+Edit} & \tablesubhead{+Ref.} & \tablesubhead{Edit $\Delta$} & \tablesubhead{Ref. $\Delta$} \\
            \midrule
            \rowcolor{white}
            \opground                 & 0.673 & 0.773 & 0.758 & \posresult{+0.100} & \posresult{+0.085} \\
            \rowcolor{tableStripe}
            \opstructured             & 0.473 & 0.404 & 0.556 & \negresult{-0.069} & \posresult{+0.083} \\
            \rowcolor{white}
            \opcue\ injection         & 0.195 & 0.365 & 0.520 & \posresult{+0.170} & \posresult{+0.325} \\
            \rowcolor{tableStripe}
            \opstate\ realization     & 0.283 & 0.580 & 0.813 & \posresult{+0.297} & \posresult{+0.530} \\
            \bottomrule
            \end{tabular}
        }
    \end{minipage}\hfill
    \begin{minipage}[c]{0.35\textwidth}
        \centering
        \includegraphics[width=\linewidth]{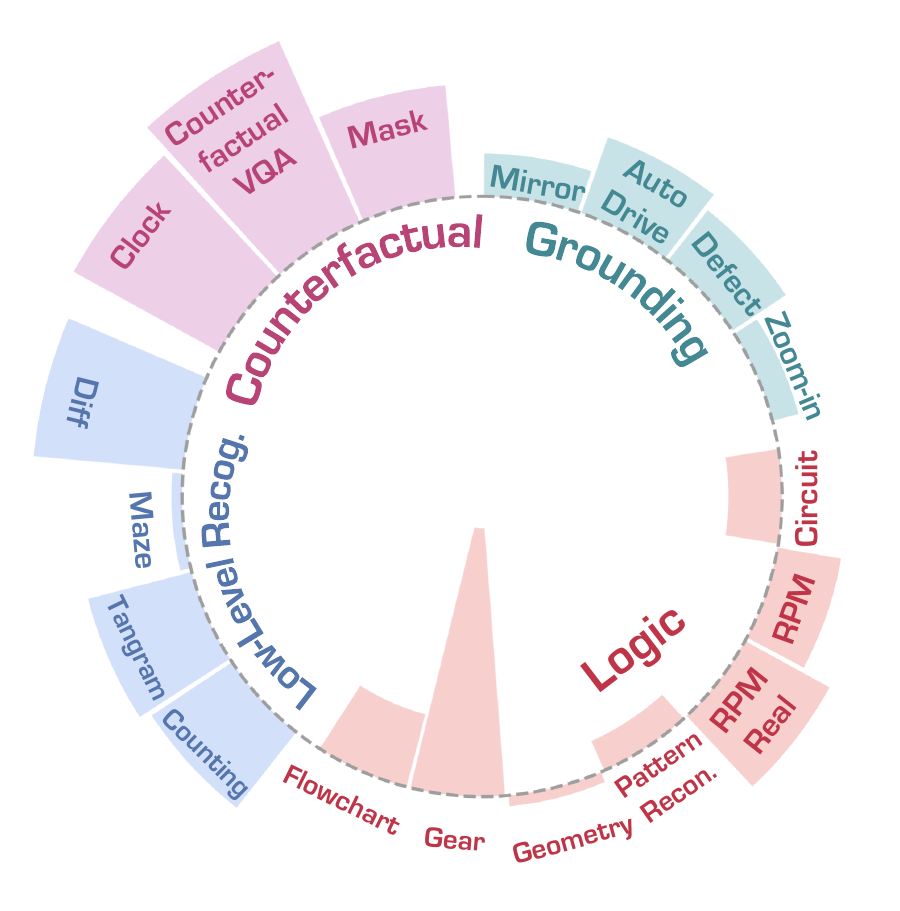}
        \captionof{figure}{\textbf{Task-Pool Composition under the Original Four Display Labels.}}
        \label{fig:sub_tasks_stat}
    \end{minipage}
    \vspace{-1em}
\end{table*}

%% file: tables/main_results_compare.tex
\begin{table*}[t]
    \centering
    \caption{\textbf{Consolidated Evaluation on the Selected Positive-Task Subset.} Left: complete within-family MLLM--editor pipelines. Right: editors compared with Qwen3-VL fixed. $\Delta$ is the absolute change in mean task score, reported in points; the Qwen pipeline's $+10.2$ points correspond to $+29.7\%$ relative to its direct score. Results characterize this selected subset rather than a general editor leaderboard.}
    \label{tab:main_results_compare}
    \footnotesize
    \setlength{\tabcolsep}{3.5pt}
    \renewcommand{\arraystretch}{1.20}
    \begin{minipage}[t]{0.51\textwidth}
        \centering
        \begin{tabular}{@{}lrrrr@{}}
        \multicolumn{5}{c}{\tablehead{(a) Complete within-family pipelines}} \\
        \toprule
        \tablesubhead{Pipeline} & \tablesubhead{Direct} & \tablesubhead{Edit} & \tablesubhead{Ref.} & \tablesubhead{$\Delta$ (pt.)} \\
        \rowcolor{tableHighlight}
        \midrule
        \textbf{Qwen3-VL + Qwen-Image-Edit} & 0.343 & \textbf{0.445} & 0.558 & \posresult{+10.2} \\
        \rowcolor{white}
        Seed-2.0 + Seedream-4.5    & 0.505 & 0.540 & 0.650 & \posresult{+3.5} \\
        \rowcolor{tableStripe}
        GPT-5 + GPT-Image-1.5      & 0.475 & 0.425 & 0.635 & \negresult{-5.0} \\
        \rowcolor{white}
        Gemini-3 + Nano Banana 2   & 0.580 & 0.625 & 0.650 & \posresult{+4.5} \\
        \bottomrule
        \end{tabular}
    \end{minipage}\hfill
    \begin{minipage}[t]{0.45\textwidth}
        \centering
        \begin{tabular}{@{}lrr@{}}
        \multicolumn{3}{c}{\tablehead{(b) Fixed Qwen3-VL reasoner}} \\
        \toprule
        \tablesubhead{Editor} & \tablesubhead{Edit score} & \tablesubhead{$\Delta$ (pt.)} \\
        \midrule
        \rowcolor{tableHighlight}
        \textbf{Qwen-Image-Edit (ours)} & \textbf{0.445} & \posresult{+10.2} \\
        \rowcolor{white}
        FLUX.2 Klein           & 0.380 & \posresult{+3.7} \\
        \rowcolor{tableStripe}
        LongCat-Image-Edit     & 0.365 & \posresult{+2.2} \\
        \rowcolor{white}
        GPT-Image-1.5          & 0.430 & \posresult{+8.7} \\
        \rowcolor{tableStripe}
        Seedream-4.5           & 0.296 & \negresult{-4.7} \\
        \rowcolor{white}
        Nano Banana 2          & 0.370 & \posresult{+2.7} \\
        \bottomrule
        \end{tabular}
    \end{minipage}
    \vspace{-1em}
\end{table*}

%% file: figures/overall_compare.tex
\begin{figure*}[htbp]
    \centering
    \includegraphics[width=0.9\textwidth]{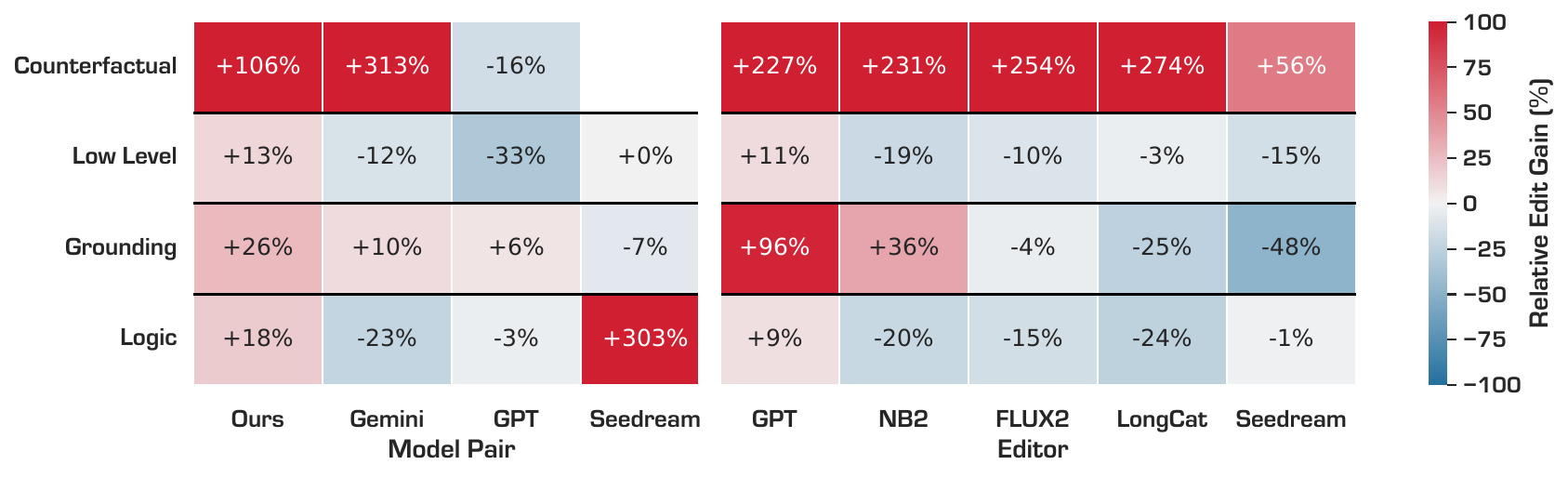}
    \caption{\textbf{Relative Edit Gain by Task Operation and Model/Editor Combination.} Large percentages can arise from small direct baselines; Table~\ref{tab:main_results_compare} reports the corresponding absolute aggregate deltas.}
    \label{fig:overall_compare}
\end{figure*}

%% file: figures/goodcases.tex
\begin{figure*}[htbp]
    \centering
    \includegraphics[width=\textwidth]{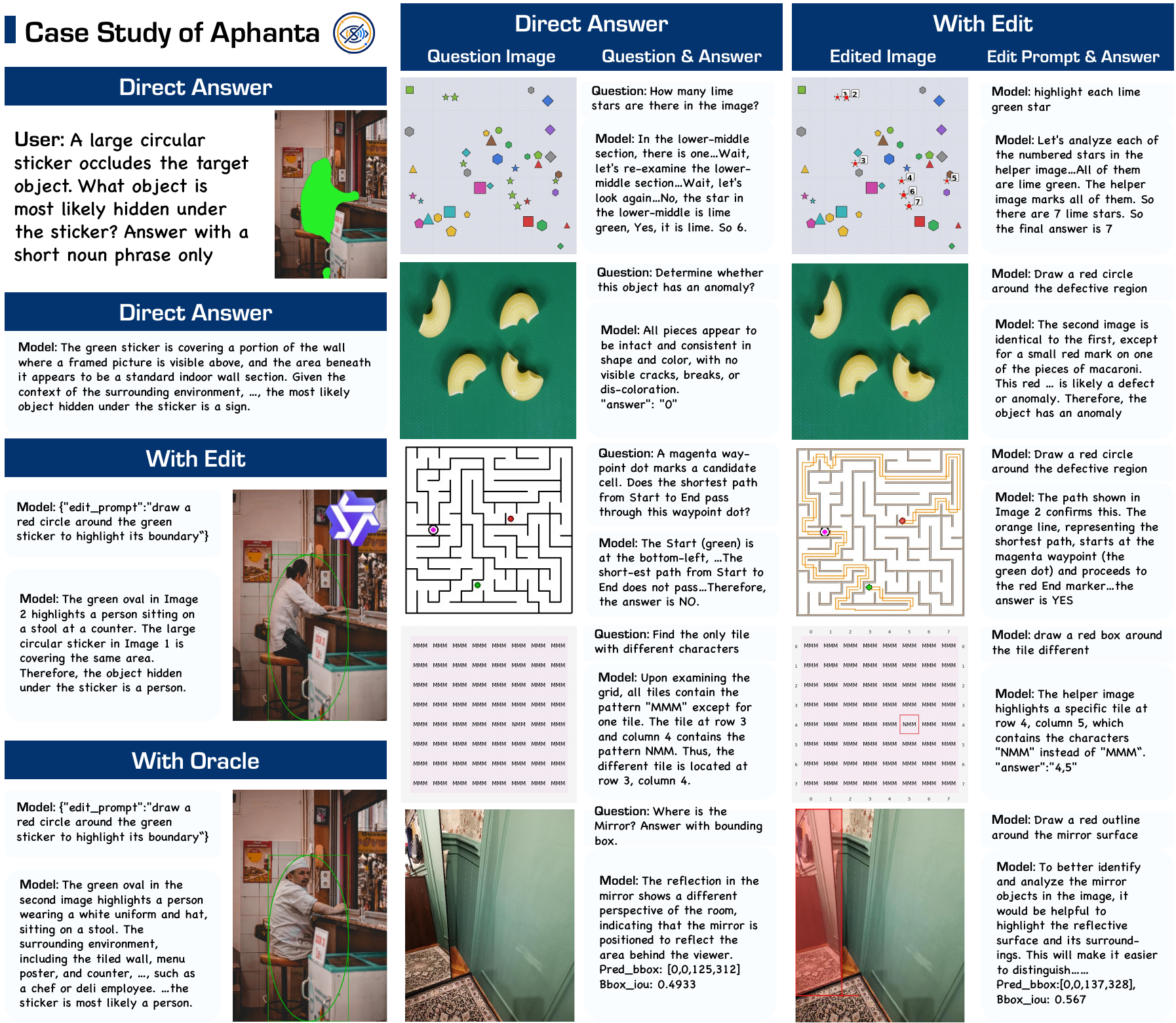}
    \caption{\textbf{Positive Cases from the Qwen3-VL and Qwen-Image-Edit Pipeline.} Each row compares direct reasoning, the actual edited intermediate, and the programmatically constructed reference condition.}
    \label{fig:goodcases}
\end{figure*}

%% file: sections/conclusion.tex
\section{Conclusion}
\label{sec:conclusion}

We introduced Aphanta, a task-discovery and closed-loop diagnostic framework for studying image-edited intermediates in multimodal reasoning. By comparing direct, actual-edit, and idealized-reference conditions across 20 candidate tasks, Aphanta separates potential visual headroom from the practical utility of current editor-in-the-loop pipelines.

The resulting map is task-conditioned. Current instruction editors are most useful for grounding, perceptual cue injection, and counterfactual state realization; they are less reliable when an intermediate must preserve exact symbols, relations, or topology. This boundary is neither a universal verdict on diffusion nor a claim that every returned image is causally used. Rather, it motivates a system design in which visual assistance is selected by task, verified after generation, and rejected or replaced when its evidence is unreliable. Aphanta provides the task pool, diagnostic protocol, and empirical baseline needed to study that broader problem.

%% file: sections/appendix.tex
\section{Experimental Details}
\label{sec:exp_details}

\subsection{Experimental Setup}

The task-development agent uses GPT-5.3-Codex with tools for dataset search and download, web retrieval, procedural construction, and Qwen-Image-Edit LoRA training on nodes with 8 NVIDIA H100 GPUs. This agent proposes and implements task assets; all quantitative scores are produced by the inference pipelines described in the main paper rather than by the development agent.

Most Qwen-Image-Edit LoRA hyperparameters are fixed across tasks. We use AdamW with learning rate $1\times10^{-4}$ and LoRA rank 32. Training length ranges from 20k to 50k steps according to data volume. At inference time, the pipeline may edit iteratively and averages approximately 1.7 editor calls per evaluated sample.

\subsection{Agent Implementation Details}

\paragraph{Agent role and execution discipline.}
The development agent is an implementation assistant for task discovery and asset construction, not the measured final reasoner. It can search for candidate data, write procedural generators, prepare reference renderers, create training and evaluation scripts, and launch jobs. The measured A/B/C scores are produced by fixed evaluation pipelines after the task assets have been constructed. To avoid uncontrolled progression, each candidate task is tracked through an ordered state machine with five phases: protocol design, preliminary validation, data collection, training handoff, and evaluation. The controller records a run id, current phase, timestamped reports, artifacts, and whether the task is finished. The data-collection and training phases have explicit gates, so the agent cannot mark them complete without a recorded approval note. This operational detail is used to keep exploratory automation from silently skipping human review.

\paragraph{Milestone reports and negative records.}
Every phase writes a timestamped report before the next phase is entered. These reports record the task hypothesis, data source, generation or filtering rules, training assets, evaluation scripts, and observed failure modes. Stopped tasks are kept in the same reporting system as successful tasks. Common stopping reasons include saturated direct baselines, negligible reference headroom, unstable synthetic data, unavailable editor outputs, and exact symbolic or topological transformations that the tested editors did not realize reliably. This is why the audit table in the main paper contains failed and stopped tasks rather than only positive demonstrations.

\paragraph{Task packet abstraction.}
Each candidate is implemented as a task packet with a common interface. A packet contains a data source or generator, a question template, an answer normalizer, an edit-instruction template or planner prompt, a reference-intermediate constructor when feasible, and a metric. The training rows for Qwen-Image-Edit are normalized to the CSV fields \texttt{edit\_image}, \texttt{image}, and \texttt{prompt}: the first image is the source, the second is the target edited image, and \texttt{prompt} is the instruction used for editor supervision. Evaluation rows are normalized to \texttt{sample\_id}, \texttt{task\_name}, \texttt{split}, \texttt{init\_image}, \texttt{question}, \texttt{answer\_gt}, \texttt{gt\_images}, and \texttt{gt\_edit\_prompts}. The final two fields store the programmatic reference image path(s) and their intended edit instruction(s), enabling the same evaluator to run Direct, Actual Edit, and Reference without task-specific glue code. Source pipeline and source file columns are appended when constructing joint datasets so that every row remains traceable.

\paragraph{Reference construction.}
Reference intermediates are constructed to express the intended visual state, not to provide an answer in text. Depending on the task, the renderer may draw bounding boxes, mark centers, add sequential indices, highlight defects, remove a specified object, complete a masked region, rotate or extend clock hands, fill a Raven matrix cell, or render a final tangram state. The reference condition is isolated from the actual-edit condition: Actual Edit receives only the editor output, while Reference receives the programmatically built target intermediate. Therefore a large +Reference gain with a weak +Edit gain indicates a realization gap rather than evidence that the task itself lacks visual headroom.

\paragraph{Joint data construction.}
The implementation includes a catalog of per-task training and evaluation paths. A joint dataset builder reads this catalog and first filters training rows to the \texttt{edit\_image,image,prompt} schema, producing a strict-schema raw export with 186,548 rows and source-tracing columns. A subsequent balanced export resamples at the pipeline level: low-resource tasks are upsampled by repeat-and-sample, high-resource tasks are downsampled, and exact-target tasks are kept unchanged. The balanced export uses a 10k-row target per pipeline, constrained by the 8k--12k policy, and contains 190,000 rows from 19 training pipelines. For evaluation, the builder selects one primary evaluation file per completed pipeline, preferring exactly 200 examples and otherwise using the test split closest to that size. The full joint evaluation set contains 2,028 rows, and the lite evaluation set contains 220 rows by sampling 20 examples from each of 11 completed pipelines. The full set is smaller than 11$\times$200 because the real Raven split contributes 28 standard test examples. Figure~\ref{fig:appendix_joint_dataset_composition} visualizes both the balancing actions and the evaluation manifest.

\input{figures/appendix_joint_dataset_composition}

\paragraph{Actual-edit inference chain.}
The unified evaluator uses a single entry point for all tested model combinations. The chain is fixed as
\[
\begin{aligned}
(x,q) &\xrightarrow{\mathrm{MLLM}_{\mathrm{plan}}} \{e_t\}_{t=1}^{T} \\
&\xrightarrow{\mathrm{Editor}} \{\tilde{x}_t\}_{t=1}^{T}
\xrightarrow{\mathrm{MLLM}_{\mathrm{answer}}} \hat{y}.
\end{aligned}
\]
The same MLLM family is used for planning and answering within a complete pipeline. The planner receives the original image and question and is required to output JSON containing only \texttt{edit\_prompts}; it is explicitly instructed not to answer at this stage. The default maximum number of planned edit steps is two in the joint evaluator, and each edited image is fed back with the original image for the final answer. For the Reference condition, the evaluator bypasses the editor and asks the same MLLM to answer from the original image plus the constructed reference image. This implements the A/B/C comparison under one scoring interface.

\paragraph{Planner prompt templates.}
The planner prompt contains in-context examples distilled from successful task packets. The examples cover traffic-sign boxes, Raven matrix completion, clock-hand rotation and completion, object removal, occlusion reconstruction, dense counting marks, tangram final-state rendering, industrial-defect highlighting, dense difference boxing, blackout/zoom-in variants, and mirror localization. Prompts are kept as short English imperative instructions because the editing backends respond most consistently to commands such as ``draw a red bounding box,'' ``remove the specified object,'' or ``highlight the anomalous region.'' If the planner emits malformed JSON, the evaluator first retries with a stricter JSON-only prompt and then falls back to a coarse task-dependent instruction, such as marking targets for counting or drawing a box around the most relevant region. The fallback keeps the pipeline executable but is logged as a planner-format failure.

\paragraph{Editor backends.}
The same evaluator supports both API-based editor combinations and local Qwen-Image-Edit. The API path covers GPT-Image-1.5, Gemini image editing, FLUX.2 Klein, LongCat-Image-Edit, Seedream-style backends when configured, and related editor services. The local Qwen path can run the base editor or load a LoRA; when given a directory, the runner selects the latest \texttt{step-*.safetensors} checkpoint. Local Qwen-Image-Edit inference uses deterministic seeds for reproducibility and writes edited images to per-sample folders. The implementation caches existing step images, so rerunning a failed metric pass does not necessarily regenerate image edits.

\paragraph{Prediction traces and metrics.}
Each evaluated sample writes a trace directory containing the copied original image, every edited helper image, sidecar JSON for each step, and a final trace JSON. The trace records the planner output, edit prompts, editor status, per-step VLM answer, final Actual Edit answer, Reference answer, and baseline Direct answer. Batch outputs include \texttt{predictions.jsonl}, \texttt{metrics.json}, \texttt{failure\_cases.jsonl}, and a viewer manifest. Accuracy tasks use normalized exact match after extracting JSON answers or common boxed-answer formats. Localization tasks use their task-specific IoU scorer. The shared summary reports Direct accuracy, Actual Edit accuracy, Reference accuracy, planner success rate, average number of planned steps, and editor success rate, while task-specific tables retain the metric appropriate to each task. Figure~\ref{fig:appendix_inference_artifacts} summarizes the branch structure and the logged artifacts.

\input{figures/appendix_inference_artifacts}

\paragraph{Human review and leakage checks.}
Human review is applied before a task enters large-scale construction and again before training handoff. Reviewers check whether the task requires the intended visual operation, whether the answer can be measured automatically, whether the reference image accidentally reveals more than the intended intermediate, and whether training and evaluation splits are separated. For generated tasks, seeds and generation parameters are kept with the task assets; for dataset-derived tasks, manifests record original paths and filtering decisions. These checks do not eliminate all task-design bias, but they reduce the chance that the measured gain comes from a malformed prompt, an ambiguous label, or a reference image that directly encodes the answer. Figure~\ref{fig:appendix_task_diagnostics} and Table~\ref{tab:edit_prompt_families} provide compact audit views of task outcomes and planner-instruction coverage.

\input{figures/appendix_task_diagnostics}

\input{tables/edit_prompt_families}

\subsection{Data Sources of Subtasks}

The data sources for the evaluated subtasks are categorized as follows:
\begin{itemize}
    \item \textbf{Open-Source Datasets}: Utilized in Task 1 \cite{yang2019my}, Task 2 \cite{ikenna113_circuitvqadesc2}, Task 3 \cite{corbiere2025drivingvqa}, Task 8 \cite{zhang2019raven}, Task 11 \cite{shi2025mathcanvasintrinsicvisualchainofthought}, Task 13 \cite{vissim_tangram_puzzle}, Task 15 \cite{zou2022spot}, and Task 16 \cite{kingsoft_qzhou_flowchart_qa}. 
    \item \textbf{In-House Edited Datasets}: Tasks 8, 9, 17, and 19 are constructed using rule-based editing pipelines applied to an in-house editing dataset.
    \item \textbf{Procedural Generation}: All remaining tasks are synthesized entirely via rule-based programmatic generation.
\end{itemize}

Beyond that, part of Zebra-CoT \cite{li2026zebracot} is used as examples in agent context, and phase-2 validation.

\subsection{Detailed Information of Tasks}

Table~\ref{tab:appendix_subtasks_full} lists the metric, training volume, preliminary diagnosis, and final-stage result when available for all 20 tasks. ``Reference'' denotes a programmatically constructed target intermediate and is used only for diagnosis.

\input{tables/subtasks_app}

\paragraph{Transfer to Existing Benchmarks.}
We additionally evaluate on BabyVision and MIRA, which were not selected by the Aphanta discovery loop, to test transfer beyond the constructed task pool.
\input{tables/benchmark_transfer_compare}
\noindent Actual edits reduce accuracy for all three tested pipelines on BabyVision. On MIRA, only Gemini-3-Flash + NB2 has a small positive delta; all three reference conditions exceed direct reasoning. These results identify reference headroom but weak practical realization under the tested pipelines, consistent with the distinction between $\Delta_{\mathrm{ref}}$ and $\Delta_{\mathrm{edit}}$ in the main paper.

%% file: figures/appendix_joint_dataset_composition.tex
\begin{figure*}[t]
\centering
\begin{tikzpicture}[x=1cm,y=1cm]
    \node[font=\bfseries\small, anchor=west] at (0,5.55) {(a) Balanced editor-training export};
    \def\trainw{5.55}
    \def\maxtrain{35}
    \newcommand{\trainx}[1]{((#1)/\maxtrain*\trainw)}
    \newcommand{\trainbar}[4]{%
        \pgfmathsetmacro{\bw}{\trainx{#3}}
        \node[font=\tiny, anchor=east] at (-0.08,#1) {#2};
        \draw[fill=tableMuted!11, draw=none] (0,#1-0.047) rectangle (\trainw,#1+0.047);
        \draw[fill=#4!78, draw=none] (0,#1-0.047) rectangle (\bw,#1+0.047);
    }
    \pgfmathsetmacro{\targetx}{\trainx{10}}
    \draw[tableMuted!35] (0,0.32) -- (0,5.02);
    \foreach \x/\lab in {0/0,10/10k,20/20k,30/30k} {
        \pgfmathsetmacro{\tx}{\trainx{\x}}
        \draw[tableMuted!25] (\tx,0.32) -- (\tx,5.02);
        \node[font=\tiny, text=tableMuted] at (\tx,0.08) {\lab};
    }
    \draw[tableHeader, dashed, line width=0.55pt] (\targetx,0.30) -- (\targetx,5.06);
    \node[font=\tiny, text=tableHeader, anchor=west] at (\targetx+0.06,5.15) {10k target};
    \trainbar{4.86}{P5}{4.018}{tablePositive}
    \trainbar{4.62}{P6}{34.500}{tableNegative}
    \trainbar{4.38}{P7}{2.303}{tablePositive}
    \trainbar{4.14}{P8}{11.000}{tableNegative}
    \trainbar{3.90}{P8.1}{2.800}{tablePositive}
    \trainbar{3.66}{P9}{20.000}{tableNegative}
    \trainbar{3.42}{P10}{10.200}{tableNegative}
    \trainbar{3.18}{P11}{2.200}{tablePositive}
    \trainbar{2.94}{P12}{9.377}{tablePositive}
    \trainbar{2.70}{P13}{1.500}{tablePositive}
    \trainbar{2.46}{P14}{8.617}{tablePositive}
    \trainbar{2.22}{P15}{10.000}{tableHeader}
    \trainbar{1.98}{P16}{2.873}{tablePositive}
    \trainbar{1.74}{P18}{2.400}{tablePositive}
    \trainbar{1.50}{P19}{31.610}{tableNegative}
    \trainbar{1.26}{P20}{3.198}{tablePositive}
    \trainbar{1.02}{P21}{10.000}{tableHeader}
    \trainbar{0.78}{A3}{10.200}{tableNegative}
    \trainbar{0.54}{A4}{9.752}{tablePositive}
    \node[font=\scriptsize, text=tableMuted] at (2.78,-0.25) {source rows before balancing};
    \draw[fill=tablePositive!78, draw=none] (0,5.28) rectangle (0.22,5.40);
    \node[font=\tiny, anchor=west] at (0.28,5.34) {upsample};
    \draw[fill=tableNegative!78, draw=none] (1.33,5.28) rectangle (1.55,5.40);
    \node[font=\tiny, anchor=west] at (1.61,5.34) {downsample};
    \draw[fill=tableHeader!78, draw=none] (2.90,5.28) rectangle (3.12,5.40);
    \node[font=\tiny, anchor=west] at (3.18,5.34) {keep};
    \node[font=\tiny, align=left, text=tableMuted, anchor=west] at (3.98,5.33) {$190{,}000$ rows\\$19\times 10k$};

    \begin{scope}[shift={(8.15,0)}]
        \node[font=\bfseries\small, anchor=west] at (0,5.55) {(b) Completed-pipeline evaluation manifest};
        \def\evalw{5.35}
        \def\maxeval{2}
        \newcommand{\evalx}[1]{((#1)/\maxeval*\evalw)}
        \newcommand{\evalbar}[4]{%
            \pgfmathsetmacro{\ew}{\evalx{#3}}
            \node[font=\tiny, anchor=east] at (-0.08,#1) {#2};
            \draw[fill=tableMuted!11, draw=none] (0,#1-0.075) rectangle (\evalw,#1+0.075);
            \draw[fill=#4!78, draw=none] (0,#1-0.075) rectangle (\ew,#1+0.075);
            \draw[black!70, line width=0.35pt] ({\evalx{0.2}},#1-0.105) -- ({\evalx{0.2}},#1+0.105);
        }
        \draw[tableMuted!35] (0,1.35) -- (0,5.00);
        \foreach \x/\lab in {0/0,0.5/50,1/100,1.5/150,2/200} {
            \pgfmathsetmacro{\ex}{\evalx{\x}}
            \draw[tableMuted!25] (\ex,1.35) -- (\ex,5.00);
            \node[font=\tiny, text=tableMuted] at (\ex,1.08) {\lab};
        }
        \evalbar{4.78}{P5}{2}{colorGrounding}
        \evalbar{4.46}{P7}{2}{colorGrounding}
        \evalbar{4.14}{P8}{2}{colorLogic}
        \evalbar{3.82}{P8.1}{0.28}{colorLogic}
        \evalbar{3.50}{P10}{2}{colorEditing}
        \evalbar{3.18}{P11}{2}{colorEditing}
        \evalbar{2.86}{P12}{2}{colorEditing}
        \evalbar{2.54}{P15}{2}{colorLowLevel}
        \evalbar{2.22}{P16}{2}{colorLowLevel}
        \evalbar{1.90}{P18}{2}{colorGrounding}
        \evalbar{1.58}{A3}{2}{colorLowLevel}
        \node[font=\scriptsize, text=tableMuted] at (2.68,0.75) {full-eval rows per pipeline};
        \draw[black!70, line width=0.35pt] (0,0.34) -- (0.32,0.34);
        \node[font=\tiny, anchor=west] at (0.42,0.34) {lite sample: 20 rows each};
        \node[font=\tiny, align=left, text=tableMuted, anchor=west] at (3.45,0.34) {full: $2{,}028$ rows\\lite: $220$ rows};
        \node[font=\tiny, text=colorGrounding, anchor=west] at (0,5.25) {Grounding};
        \node[font=\tiny, text=colorEditing, anchor=west] at (1.24,5.25) {State};
        \node[font=\tiny, text=colorLowLevel, anchor=west] at (2.00,5.25) {Cue};
        \node[font=\tiny, text=colorLogic, anchor=west] at (2.62,5.25) {Structured};
    \end{scope}
\end{tikzpicture}
\caption{\textbf{Joint Data Construction Diagnostics.} Panel (a) visualizes the balanced training export produced from the catalog-level training manifest: source rows are shown before resampling, and every pipeline is resampled to 10k rows. Panel (b) visualizes the completed-pipeline evaluation manifest. All lite splits sample 20 rows per pipeline; the full split uses 200 rows when available, except the real Raven split (P8.1), which contributes 28 standard test examples.}
\label{fig:appendix_joint_dataset_composition}
\end{figure*}
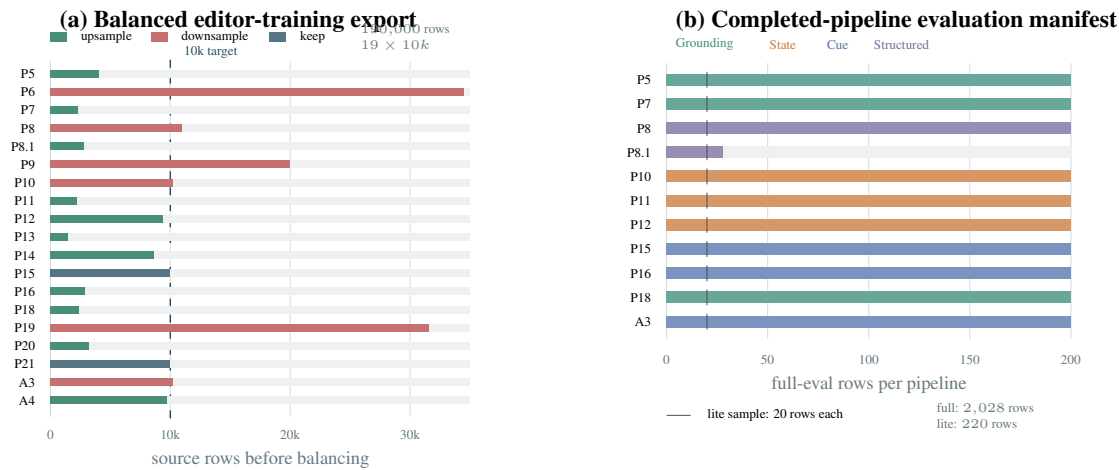

%% file: figures/appendix_inference_artifacts.tex
\begin{figure*}[t]
\centering
\begin{tikzpicture}[x=1cm,y=1cm]
    \tikzset{
        afbox/.style={draw=tableMuted!48, fill=tableStripe, rounded corners=1.5pt, align=center, inner sep=3.2pt, font=\scriptsize},
        afhead/.style={draw=tableHeader!65, fill=tableSubheader, rounded corners=1.5pt, align=center, inner sep=3.2pt, font=\scriptsize\bfseries},
        afout/.style={draw=tablePositive!50, fill=tablePositive!8, rounded corners=1.5pt, align=center, inner sep=3.2pt, font=\scriptsize}
    }
    \node[afhead, minimum width=3.15cm] (row) at (0,3.75) {Evaluation row\\\texttt{init\_image}, \texttt{question}\\\texttt{answer\_gt}, \texttt{gt\_images}};
    \node[afbox, minimum width=3.05cm] (direct) at (-5.3,2.25) {A: Direct\\original image + question\\answer MLLM};
    \node[afbox, minimum width=3.15cm] (plan) at (0,2.25) {Planner MLLM\\JSON only:\\\texttt{edit\_prompts}};
    \node[afbox, minimum width=3.05cm] (ref) at (5.3,2.25) {C: Reference\\original + constructed\\reference image(s)};
    \node[afbox, minimum width=3.15cm] (edit) at (0,0.98) {Editor backend\\API editor or local\\Qwen-Image-Edit LoRA};
    \node[afbox, minimum width=3.15cm] (answer) at (0,-0.22) {B: Actual Edit\\original + edited step image(s)\\answer MLLM};
    \node[afout, minimum width=3.10cm] (pred) at (-5.3,-1.70) {\texttt{predictions.jsonl}\\A/B/C answers\\trace path};
    \node[afout, minimum width=3.10cm] (metrics) at (0,-1.70) {\texttt{metrics.json}\\accuracy or IoU\\planner/editor rates};
    \node[afout, minimum width=3.10cm] (viewer) at (5.3,-1.70) {\texttt{failure\_cases.jsonl}\\\texttt{viewer\_manifest.json}\\per-sample traces};

    \draw[->, tableMuted!70] (row) -- (direct);
    \draw[->, tableMuted!70] (row) -- (plan);
    \draw[->, tableMuted!70] (row) -- (ref);
    \draw[->, tableMuted!70] (plan) -- (edit);
    \draw[->, tableMuted!70] (edit) -- (answer);
    \draw[->, tableMuted!70] (direct) -- (pred);
    \draw[->, tableMuted!70] (answer) -- (metrics);
    \draw[->, tableMuted!70] (ref) -- (viewer);
    \draw[->, tableMuted!45] (pred) -- (metrics);
    \draw[->, tableMuted!45] (metrics) -- (viewer);

    \node[font=\tiny, text=tableMuted, align=center] at (-2.72,3.26) {baseline branch};
    \node[font=\tiny, text=tableMuted, align=center] at (2.64,3.26) {diagnostic branch};
    \node[font=\tiny, text=tableMuted, align=left, anchor=west] at (1.78,1.58) {max two planned edit steps\\in the joint evaluator};
    \node[font=\tiny, text=tableMuted, align=center] at (0,-2.50) {Each sample keeps copied inputs, edited images, sidecar step JSON, and a final trace JSON.};
\end{tikzpicture}
\caption{\textbf{Unified Inference and Artifact Flow.} The same normalized evaluation row feeds Direct, Actual Edit, and Reference branches. The Actual Edit branch first plans short edit instructions, then calls an editor backend and re-answers from the edited helper image(s). The evaluator writes both metric-level summaries and per-sample traces, which makes planner failures, editor failures, and answer errors auditable after a run.}
\label{fig:appendix_inference_artifacts}
\end{figure*}
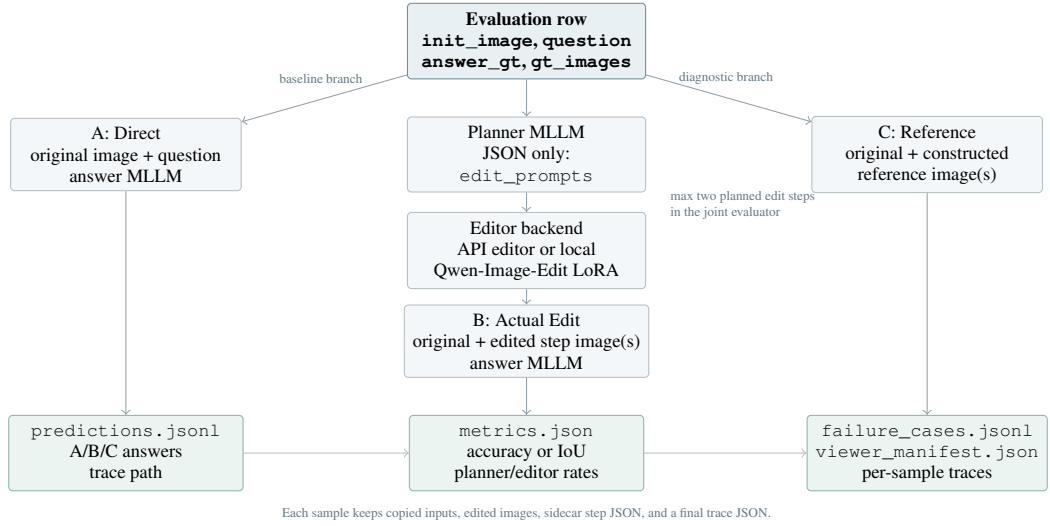

%% file: figures/appendix_task_diagnostics.tex
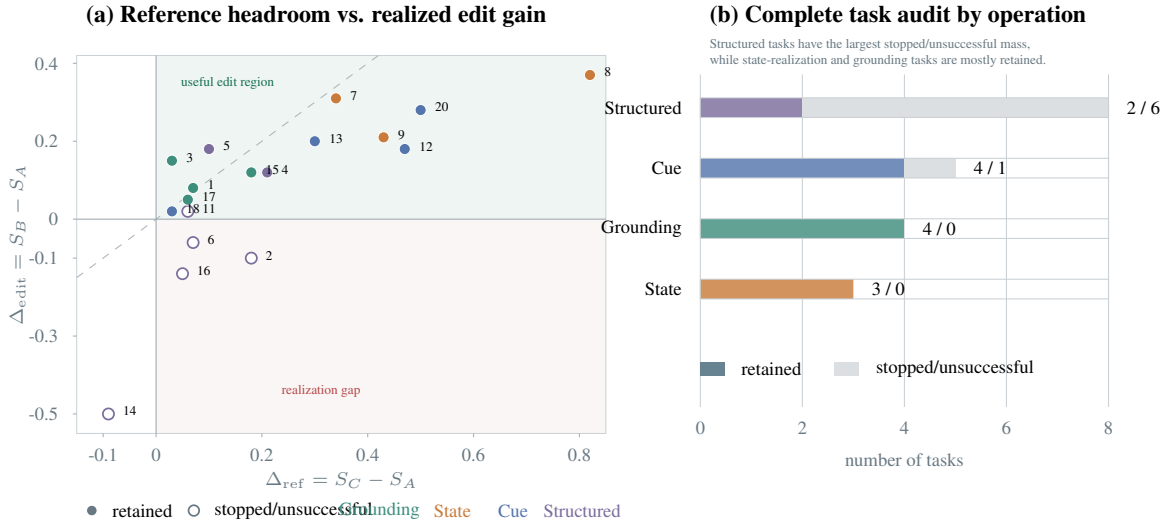
\begin{figure*}[t]
\centering
\begin{tikzpicture}[x=1cm,y=1cm]
    \def\xmin{-0.15}
    \def\xmax{0.85}
    \def\ymin{-0.55}
    \def\ymax{0.42}
    \def\plotw{7.0}
    \def\ploth{5.0}

    \newcommand{\mapx}[1]{(((#1)-(\xmin))/((\xmax)-(\xmin))*\plotw)}
    \newcommand{\mapy}[1]{(((#1)-(\ymin))/((\ymax)-(\ymin))*\ploth)}
    \newcommand{\tickx}[2]{%
        \pgfmathsetmacro{\tx}{\mapx{#1}}
        \draw[tableMuted!60] (\tx,0) -- (\tx,-0.05);
        \node[font=\scriptsize, text=tableMuted] at (\tx,-0.28) {#2};
    }
    \newcommand{\ticky}[2]{%
        \pgfmathsetmacro{\ty}{\mapy{#1}}
        \draw[tableMuted!60] (0,\ty) -- (-0.05,\ty);
        \node[font=\scriptsize, text=tableMuted, anchor=east] at (-0.12,\ty) {#2};
    }
    \newcommand{\passpt}[4]{%
        \pgfmathsetmacro{\px}{\mapx{#1}}
        \pgfmathsetmacro{\py}{\mapy{#2}}
        \node[circle, fill=#3, draw=white, line width=0.25pt, minimum size=4.2pt, inner sep=0pt] at (\px,\py) {};
        \node[font=\tiny, text=black, anchor=west] at (\px+0.07,\py+0.04) {#4};
    }
    \newcommand{\failpt}[4]{%
        \pgfmathsetmacro{\px}{\mapx{#1}}
        \pgfmathsetmacro{\py}{\mapy{#2}}
        \node[circle, fill=white, draw=#3, line width=0.65pt, minimum size=4.2pt, inner sep=0pt] at (\px,\py) {};
        \node[font=\tiny, text=black, anchor=west] at (\px+0.07,\py+0.04) {#4};
    }

    \node[font=\bfseries\small, anchor=west] at (0,5.52) {(a) Reference headroom vs. realized edit gain};
    \pgfmathsetmacro{\xzero}{\mapx{0}}
    \pgfmathsetmacro{\yzero}{\mapy{0}}
    \pgfmathsetmacro{\diagxA}{\mapx{-0.15}}
    \pgfmathsetmacro{\diagyA}{\mapy{-0.15}}
    \pgfmathsetmacro{\diagxB}{\mapx{0.42}}
    \pgfmathsetmacro{\diagyB}{\mapy{0.42}}
    \pgfmathsetmacro{\labelxA}{\mapx{0.03}}
    \pgfmathsetmacro{\labelyA}{\mapy{0.35}}
    \pgfmathsetmacro{\labelxB}{\mapx{0.22}}
    \pgfmathsetmacro{\labelyB}{\mapy{-0.44}}
    \fill[tablePositive!7] (\xzero,\yzero) rectangle (\plotw,\ploth);
    \fill[tableNegative!5] (\xzero,0) rectangle (\plotw,\yzero);
    \draw[tableMuted!35, line width=0.35pt] (0,0) rectangle (\plotw,\ploth);
    \foreach \x/\lab in {-0.1/-0.1,0/0,0.2/0.2,0.4/0.4,0.6/0.6,0.8/0.8} {\tickx{\x}{\lab}}
    \foreach \y/\lab in {-0.5/-0.5,-0.3/-0.3,-0.1/-0.1,0/0,0.2/0.2,0.4/0.4} {\ticky{\y}{\lab}}
    \draw[tableMuted!55, dashed] (\diagxA,\diagyA) -- (\diagxB,\diagyB);
    \draw[tableMuted!65] (\xzero,0) -- (\xzero,\ploth);
    \draw[tableMuted!65] (0,\yzero) -- (\plotw,\yzero);
    \node[font=\scriptsize, text=tableMuted] at (3.5,-0.62) {$\Delta_{\mathrm{ref}}=S_C-S_A$};
    \node[font=\scriptsize, text=tableMuted, rotate=90] at (-0.78,2.5) {$\Delta_{\mathrm{edit}}=S_B-S_A$};
    \node[font=\tiny, text=tablePositive, anchor=west] at (\labelxA,\labelyA) {useful edit region};
    \node[font=\tiny, text=tableNegative, anchor=west] at (\labelxB,\labelyB) {realization gap};

    \passpt{0.07}{0.08}{colorGrounding}{1}
    \failpt{0.18}{-0.10}{colorLogic}{2}
    \passpt{0.03}{0.15}{colorGrounding}{3}
    \passpt{0.21}{0.12}{colorLogic}{4}
    \passpt{0.10}{0.18}{colorLogic}{5}
    \failpt{0.07}{-0.06}{colorLogic}{6}
    \passpt{0.34}{0.31}{colorEditing}{7}
    \passpt{0.82}{0.37}{colorEditing}{8}
    \passpt{0.43}{0.21}{colorEditing}{9}
    \failpt{0.06}{0.02}{colorLogic}{11}
    \passpt{0.47}{0.18}{colorLowLevel}{12}
    \passpt{0.30}{0.20}{colorLowLevel}{13}
    \failpt{-0.09}{-0.50}{colorLogic}{14}
    \passpt{0.18}{0.12}{colorGrounding}{15}
    \failpt{0.05}{-0.14}{colorLogic}{16}
    \passpt{0.06}{0.05}{colorGrounding}{17}
    \passpt{0.03}{0.02}{colorLowLevel}{18}
    \passpt{0.50}{0.28}{colorLowLevel}{20}

    \node[circle, fill=tableMuted, draw=white, minimum size=4pt, inner sep=0pt] at (0.2,-1.02) {};
    \node[font=\scriptsize, anchor=west] at (0.35,-1.02) {retained};
    \node[circle, fill=white, draw=tableMuted, line width=0.65pt, minimum size=4pt, inner sep=0pt] at (1.55,-1.02) {};
    \node[font=\scriptsize, anchor=west] at (1.70,-1.02) {stopped/unsuccessful};
    \node[font=\scriptsize, text=colorGrounding, anchor=west] at (3.35,-1.02) {Grounding};
    \node[font=\scriptsize, text=colorEditing, anchor=west] at (4.60,-1.02) {State};
    \node[font=\scriptsize, text=colorLowLevel, anchor=west] at (5.45,-1.02) {Cue};
    \node[font=\scriptsize, text=colorLogic, anchor=west] at (6.05,-1.02) {Structured};

    \begin{scope}[shift={(8.25,0)}]
        \node[font=\bfseries\small, anchor=west] at (0,5.52) {(b) Complete task audit by operation};
        \def\barw{5.4}
        \def\maxn{8}
        \newcommand{\statusbar}[5]{%
            \pgfmathsetmacro{\pw}{#2/\maxn*\barw}
            \pgfmathsetmacro{\fw}{#3/\maxn*\barw}
            \node[font=\scriptsize, anchor=east] at (-0.12,#1+0.13) {#5};
            \draw[fill=#4!82, draw=none] (0,#1) rectangle (\pw,#1+0.26);
            \draw[fill=tableMuted!24, draw=none] (\pw,#1) rectangle (\pw+\fw,#1+0.26);
            \draw[tableMuted!35] (0,#1) rectangle (\barw,#1+0.26);
            \node[font=\scriptsize, anchor=west] at (\pw+\fw+0.12,#1+0.13) {#2 / #3};
        }
        \foreach \x/\lab in {0/0,2/2,4/4,6/6,8/8} {
            \pgfmathsetmacro{\gx}{\x/\maxn*\barw}
            \draw[tableMuted!35] (\gx,0.35) -- (\gx,4.75);
            \node[font=\scriptsize, text=tableMuted] at (\gx,0.08) {\lab};
        }
        \statusbar{4.18}{2}{6}{colorLogic}{Structured}
        \statusbar{3.38}{4}{1}{colorLowLevel}{Cue}
        \statusbar{2.58}{4}{0}{colorGrounding}{Grounding}
        \statusbar{1.78}{3}{0}{colorEditing}{State}
        \node[font=\scriptsize, text=tableMuted] at (2.7,-0.35) {number of tasks};
        \draw[fill=tableHeader!70, draw=none] (0,0.76) rectangle (0.32,0.94);
        \node[font=\scriptsize, anchor=west] at (0.42,0.85) {retained};
        \draw[fill=tableMuted!24, draw=none] (1.78,0.76) rectangle (2.10,0.94);
        \node[font=\scriptsize, anchor=west] at (2.20,0.85) {stopped/unsuccessful};
        \node[font=\tiny, text=tableMuted, align=left, anchor=west] at (0,5.03) {Structured tasks have the largest stopped/unsuccessful mass,\\while state-realization and grounding tasks are mostly retained.};
    \end{scope}
\end{tikzpicture}
\caption{\textbf{Diagnostic Views over the 20-Task Audit.} Panel (a) plots the reference gain against the actual edit gain for tasks with both quantities available; filled markers denote retained practical pipelines, and open markers denote stopped or unsuccessful tasks. Tasks 10 and 19 lack an actual-edit score and are omitted from the scatter but included in panel (b). Panel (b) summarizes retained versus stopped/unsuccessful outcomes by dominant operation.}
\label{fig:appendix_task_diagnostics}
\end{figure*}

%% file: tables/edit_prompt_families.tex
\begin{table*}[t]
    \centering
    \caption{\textbf{Edit-Instruction Template Families Used by the Unified Planner.} Templates are distilled from successful task packets and injected as in-context examples for planning; concrete object names, directions, angles, and target categories are filled from each sample.}
    \label{tab:edit_prompt_families}
    \footnotesize
    \setlength{\tabcolsep}{4pt}
    \renewcommand{\arraystretch}{1.16}
    \resizebox{\textwidth}{!}{%
    \begin{tabular}{p{0.18\textwidth} p{0.28\textwidth} p{0.34\textwidth} p{0.12\textwidth}}
    \toprule
    \tablehead{Template family} & \tablehead{Representative tasks} & \tablehead{Typical instruction form} & \tablehead{Planner role} \\
    \midrule
    \rowcolor{white}
    Fixed spatial annotation & Mirror localization, traffic signs, industrial defects, dense difference & Draw red boxes around the relevant region; highlight anomalous regions; mark the only differing tile. & Cue grounding \\
    \rowcolor{tableStripe}
    Parameterized local edit & Clock reasoning, object deletion, dense counting & Rotate the minute hand by a specified angle; remove \texttt{\{object\}}; mark each \texttt{\{target\}} with a red dot and index. & State update \\
    \rowcolor{white}
    Constructive final state & Synthetic and real RPM, tangram & Fill the blank matrix cell with the final pattern; generate the final tangram board after the described steps. & Visual construction \\
    \rowcolor{tableStripe}
    Isolation and close-up & Zoom-in/remake and blackout variants & Keep only the target object and black out the rest; crop the edited subject into a square close-up. & Evidence filtering \\
    \bottomrule
    \end{tabular}%
    }
\end{table*}

%% file: tables/subtasks_app.tex
\begin{table*}[thbp]
    \centering
    \caption{\textbf{Detailed Task Audit.} Preliminary and final-stage scores are reported separately. Reference intermediates are programmatically constructed diagnostic targets.}
    \label{tab:appendix_subtasks_full}
    \footnotesize
    \setlength{\tabcolsep}{3.6pt}
    \renewcommand{\arraystretch}{1.18}
    \resizebox{0.9\textwidth}{!}{%
    \begin{tabular}{c l c l c c @{\hspace{6pt}} c c c @{\hspace{6pt}} c c c}
    \toprule
    \multirow{2}[2]{*}{\tablehead{ID}} &
    \multirow{2}[2]{*}{\tablehead{Task}} &
    \multirow{2}[2]{*}{\tablehead{Status}} &
    \multirow{2}[2]{*}{\tablehead{Operation}} &
    \multirow{2}[2]{*}{\tablehead{Metric}} &
    \multirow{2}[2]{*}{\tablehead{Train $N$}} &
    \multicolumn{3}{c}{\tablehead{Preliminary diagnosis}} &
    \multicolumn{3}{c}{\tablehead{Final evaluation}} \\
    \cmidrule(lr){7-9}\cmidrule(l){10-12}
    & & & & & & \tablesubhead{Direct} & \tablesubhead{+Edit} & \tablesubhead{+Ref.} & \tablesubhead{Direct} & \tablesubhead{+Edit} & \tablesubhead{+Ref.} \\
    \midrule
    \rowcolor{white}
    1 & Where Is My Mirror & \statuspass & \opground & IOU & 3,818 & 0.53 & 0.53 & 0.50 & 0.70 & 0.78 & 0.77 \\
    \rowcolor{tableStripe}
    2 & Circuit Diagram Parsing & \statusfail & \opstructured & ACC & 30,155 & 0.70 & 0.60 & 0.88 & -- & -- & -- \\
    \rowcolor{white}
    3 & Auto. Driving Assistant & \statuspass & \opground & ACC & 2,303 & 0.45 & 0.43 & 0.63 & 0.42 & 0.57 & 0.45 \\
    \rowcolor{tableStripe}
    4 & RPM: Synthetic & \statuspass & \opstructured & ACC & 10,000 & 0.24 & 0.18 & 0.37 & 0.18 & 0.30 & 0.39 \\
    \rowcolor{white}
    5 & RPM: Real & \statuspass & \opstructured & ACC & 2,100 & 0.51 & 0.43 & 0.60 & 0.61 & 0.79 & 0.71 \\
    \rowcolor{tableStripe}
    6 & Repeated Pattern Recog. & \statusfail & \opstructured & ACC & 20,000 & 0.13 & 0.07 & 0.20 & -- & -- & -- \\
    \rowcolor{white}
    7 & Analog Clock Reasoning & \statuspass & \opstate & ACC & 10,000 & 0.46 & 0.58 & 0.79 & 0.50 & 0.81 & 0.84 \\
    \rowcolor{tableStripe}
    8 & Counterfactual: Deletion & \statuspass & \opstate & ACC & 2,200 & 0.20 & 0.50 & 0.87 & 0.10 & 0.47 & 0.92 \\
    \rowcolor{white}
    9 & What Is Behind the Mask? & \statuspass & \opstate & ACC & 9,377 & 0.08 & 0.04 & 0.21 & 0.25 & 0.46 & 0.68 \\
    \rowcolor{tableStripe}
    10 & Visual Equation Puzzle & \statusfail & \opstructured & ACC & -- & 0.85 & -- & 0.85 & -- & -- & -- \\
    \rowcolor{white}
    11 & Plane Geometry Aux. Line & \statusfail & \opstructured & ACC & 8,417 & 0.40 & 0.20 & 0.40 & 0.41 & 0.43 & 0.47 \\
    \rowcolor{tableStripe}
    12 & Dense Dot Counting & \statuspass & \opcue & ACC & 9,800 & 0.08 & 0.06 & 0.83 & 0.05 & 0.23 & 0.52 \\
    \rowcolor{white}
    13 & Tangram & \statuspass & \opcue & ACC & 2,673 & 0.40 & 0.35 & 0.35 & 0.31 & 0.51 & 0.61 \\
    \rowcolor{tableStripe}
    14 & Gear Rotation Reasoning & \statusfail & \opstructured & ACC & -- & 0.92* & 0.42* & 0.83* & -- & -- & -- \\
    \rowcolor{white}
    15 & Industrial Defect Inspect & \statuspass & \opground & ACC & 2,200 & 0.63 & 0.17 & 0.83 & 0.70 & 0.82 & 0.88 \\
    \rowcolor{tableStripe}
    16 & Flowchart Decision & \statusfail & \opstructured & ACC & 31,410 & 0.37 & 0.20 & 0.43 & 0.36 & 0.22 & 0.41 \\
    \rowcolor{white}
    17 & Zoom-in (Remake) & \statuspass & \opground & ACC & 2,998 & 0.75 & 0.80 & 0.90 & 0.87 & 0.92 & 0.93 \\
    \rowcolor{tableStripe}
    18 & Maze Solving & \statuspass & \opcue & ACC & 10,000 & -- & -- & -- & 0.24 & 0.26 & 0.27 \\
    \rowcolor{white}
    19 & Spot-the-Diff (Sparse) & \statusfail & \opcue & ACC & 7,709 & 0.53 & -- & 0.58 & -- & -- & -- \\
    \rowcolor{tableStripe}
    20 & Spot-the-Diff (Dense) & \statuspass & \opcue & ACC & 10,000 & 0.00 & 0.00 & 0.75 & 0.18 & 0.46 & 0.68 \\
    \bottomrule
    \end{tabular}%
    }
\end{table*}

%% file: tables/benchmark_transfer_compare.tex
\begin{table*}[t]
    \centering
    \caption{\textbf{Transfer to BabyVision and MIRA.} Edit $\Delta$ and Ref. $\Delta$ are absolute accuracy changes from direct reasoning; MIRA additionally provides constructed reference intermediates.}
    \label{tab:benchmark_transfer_compare}
    \footnotesize
    \renewcommand{\arraystretch}{1.20}

    \begin{minipage}[t]{0.3\textwidth}
    \centering
    \resizebox{\linewidth}{!}{
    \begin{tabular}{@{}lccc@{}}
    \multicolumn{4}{c}{\tablehead{(a) BabyVision}} \\
    \toprule
    \tablesubhead{Reasoner} & \tablesubhead{Direct} & \tablesubhead{+Edit} & \tablesubhead{Edit $\Delta$} \\
    \midrule
    \rowcolor{white}
    Qwen & 0.1958 & 0.1623 & \negresult{-0.0335} \\
    \rowcolor{tableStripe}
    GPT & 0.1804 & 0.1546 & \negresult{-0.0258} \\
    \rowcolor{white}
    Gemini & 0.3015 & 0.2319 & \negresult{-0.0696} \\
    \bottomrule
    \end{tabular}
    }
    \end{minipage}%
    \hfill
    \begin{minipage}[t]{0.45\textwidth}
    \centering
    \resizebox{\linewidth}{!}{
    \begin{tabular}{@{}lccccc@{}}
    \multicolumn{6}{c}{\tablehead{(b) MIRA}} \\
    \toprule
    \tablesubhead{Reasoner} & \tablesubhead{Direct} & \tablesubhead{+Edit} & \tablesubhead{+Ref.} & \tablesubhead{Edit $\Delta$} & \tablesubhead{Ref. $\Delta$} \\
    \midrule
    \rowcolor{white}
    Qwen & 0.1941 & 0.1664 & 0.2421 & \negresult{-0.0277} & \posresult{+0.0480} \\
    \rowcolor{tableStripe}
    GPT & 0.1832 & 0.1337 & 0.2473 & \negresult{-0.0495} & \posresult{+0.0641} \\
    \rowcolor{white}
    Gemini & 0.2033 & 0.2125 & 0.2930 & \posresult{+0.0092} & \posresult{+0.0897} \\
    \bottomrule
    \end{tabular}
    }
    \end{minipage}
\end{table*}

%% file: main.bib
@string(CVPR= {IEEE Conf. Comput. Vis. Pattern Recog.})

@string(CVPR  = {CVPR})

@article{yin2025reasonedit,
  title={ReasonEdit: Towards Reasoning-Enhanced Image Editing Models}, 
  author={Fukun Yin and Shiyu Liu and Yucheng Han and Zhibo Wang and Peng Xing and Rui Wang and Wei Cheng and Yingming Wang and Aojie Li and Zixin Yin and Pengtao Chen and Xiangyu Zhang and Daxin Jiang and Xianfang Zeng and Gang Yu},
  journal={arXiv preprint arXiv:2511.22625},
  year={2025}
}

@article{wu2025kris,
  title={KRIS-Bench: Benchmarking Next-Level Intelligent Image Editing Models},
  author={Wu, Yongliang and Li, Zonghui and Hu, Xinting and Ye, Xinyu and Zeng, Xianfang and Yu, Gang and Zhu, Wenbo and Schiele, Bernt and Yang, Ming-Hsuan and Yang, Xu},
  journal={arXiv preprint arXiv:2505.16707},
  year={2025}
}

@article{liu2025step1x-edit,
  title={Step1X-Edit: A Practical Framework for General Image Editing}, 
  author={Shiyu Liu and Yucheng Han and Peng Xing and Fukun Yin and Rui Wang and Wei Cheng and Jiaqi Liao and Yingming Wang and Honghao Fu and Chunrui Han and Guopeng Li and Yuang Peng and Quan Sun and Jingwei Wu and Yan Cai and Zheng Ge and Ranchen Ming and Lei Xia and Xianfang Zeng and Yibo Zhu and Binxing Jiao and Xiangyu Zhang and Gang Yu and Daxin Jiang},
  journal={arXiv preprint arXiv:2504.17761},
  year={2025}
}

@article{hong2025deepeyesv2,
  title={DeepEyesV2: Toward Agentic Multimodal Model},
  author={Hong, Jack and Zhao, Chenxiao and Zhu, ChengLin and Lu, Weiheng and Xu, Guohai and Yu, Xing},
  journal={arXiv preprint arXiv:2511.05271},
  year={2025}
}

@article{zheng2025deepeyes,
  title={DeepEyes: Incentivizing" Thinking with Images" via Reinforcement Learning},
  author={Zheng, Ziwei and Yang, Michael and Hong, Jack and Zhao, Chenxiao and Xu, Guohai and Yang, Le and Shen, Chao and Yu, Xing},
  journal={arXiv preprint arXiv:2505.14362},
  year={2025}
}

@article{zhang2025thyme,
  title={Thyme: Think Beyond Images},
  author={Zhang, Yi-Fan and Lu, Xingyu and Yin, Shukang and Fu, Chaoyou and Chen, Wei and Hu, Xiao and Wen, Bin and Jiang, Kaiyu and Liu, Changyi and Zhang, Tianke and others},
  journal={arXiv preprint arXiv:2508.11630},
  year={2025}
}

@article{lai2025mini-o3,
  title={Mini-o3: Scaling Up Reasoning Patterns and Interaction Turns for Visual Search},
  author={Lai, Xin and Li, Junyi and Li, Wei and Liu, Tao and Li, Tianjian and Zhao, Hengshuang},
  journal={arXiv:2509.07969},
  year={2025}
}

@inproceedings{
  li2026zebracot,
  title={Zebra-CoT: A Dataset for Interleaved Vision-Language Reasoning},
  author={Ang Li and Charles Wang and Deqing Fu and Kaiyu Yue and Zikui Cai
          and Wang Bill Zhu and Ollie Liu and Peng Guo and Willie Neiswanger
          and Furong Huang and Tom Goldstein and Micah Goldblum},
  booktitle={The Fourteenth International Conference on Learning Representations},
  year={2026},
  url={https://openreview.net/forum?id=c6XIVI3TiQ}
}

@article{duan2025codeplot,
  title={CodePlot-CoT: Mathematical Visual Reasoning by Thinking with Code-Driven Images},
  author={Duan, Chengqi and Sun, Kaiyue and Fang, Rongyao and Zhang, Manyuan and Feng, Yan and Luo, Ying and Liu, Yufang and Wang, Ke and Pei, Peng and Cai, Xunliang and others},
  journal={arXiv preprint arXiv:2510.11718},
  year={2025}
}

@article{qiao2025v,
  title={V-Thinker: Interactive Thinking with Images},
  author={Qiao, Runqi and Tan, Qiuna and Yang, Minghan and Dong, Guanting and Yang, Peiqing and Lang, Shiqiang and Wan, Enhui and Wang, Xiaowan and Xu, Yida and Yang, Lan and others},
  journal={arXiv preprint arXiv:2511.04460},
  year={2025}
}

@misc{chen2026babyvisionvisualreasoninglanguage,
      title={BabyVision: Visual Reasoning Beyond Language}, 
      author={Liang Chen and Weichu Xie and Yiyan Liang and Hongfeng He and Hans Zhao and Zhibo Yang and Zhiqi Huang and Haoning Wu and Haoyu Lu and Y. charles and Yiping Bao and Yuantao Fan and Guopeng Li and Haiyang Shen and Xuanzhong Chen and Wendong Xu and Shuzheng Si and Zefan Cai and Wenhao Chai and Ziqi Huang and Fangfu Liu and Tianyu Liu and Baobao Chang and Xiaobo Hu and Kaiyuan Chen and Yixin Ren and Yang Liu and Yuan Gong and Kuan Li},
      year={2026},
      eprint={2601.06521},
      archivePrefix={arXiv},
      primaryClass={cs.CV},
      url={https://arxiv.org/abs/2601.06521}, 
}

@misc{song2025visualpuzzlesdecouplingmultimodalreasoning,
  title={VisualPuzzles: Decoupling Multimodal Reasoning Evaluation from Domain Knowledge},
  author={Yueqi Song and Tianyue Ou and Yibo Kong and Zecheng Li and Graham Neubig and Xiang Yue},
  year={2025},
  eprint={2504.10342},
  archivePrefix={arXiv},
  primaryClass={cs.CL},
  url={https://arxiv.org/abs/2504.10342}
}

@article{xu2025visulogic,
  title={VisuLogic: A Benchmark for Evaluating Visual Reasoning in Multi-modal Large Language Models},
  author={Xu, Weiye and Wang, Jiahao and Wang, Weiyun and Chen, Zhe and Zhou, Wengang and Yang, Aijun and Lu, Lewei and Li, Houqiang and Wang, Xiaohua and Zhu, Xizhou and Wang, Wenhai and Dai, Jifeng and Zhu, Jinguo},
  journal={arXiv preprint arXiv:2504.15279},
  year={2025},
  url={https://arxiv.org/abs/2504.15279}
}

@misc{zhou2025visualizingstepreasoningmira,
      title={When Visualizing is the First Step to Reasoning: MIRA, a Benchmark for Visual Chain-of-Thought},
      author={Yiyang Zhou and Haoqin Tu and Zijun Wang and Zeyu Wang and Niklas Muennighoff and Fan Nie and Yejin Choi and James Zou and Chaorui Deng and Shen Yan and Haoqi Fan and Cihang Xie and Huaxiu Yao and Qinghao Ye},
      year={2025},
      eprint={2511.02779},
      archivePrefix={arXiv},
      primaryClass={cs.CV},
      url={https://arxiv.org/abs/2511.02779},
}

@article{he2025diffthinker,
  title={DiffThinker: Towards Generative Multimodal Reasoning with Diffusion Models},
  author={He, Zefeng and Qu, Xiaoye and Li, Yafu and Zhu, Tong and Huang, Siyuan and Cheng, Yu},
  journal={arXiv preprint arXiv:2512.24165},
  year={2025}
}

@article{dai2026endocot,
  title={EndoCoT: Scaling Endogenous Chain-of-Thought Reasoning in Diffusion Models},
  author={Dai, Xuanlang and Zhou, Yujie and Xing, Long and Bu, Jiazi and Wei, Xilin and Liu, Yuhong and Zhang, Beichen and Chen, Kai and Zang, Yuhang},
  journal={arXiv preprint arXiv:2603.12252},
  year={2026}
}

@article{chern2025thinkingwithgeneratedimages,
  title={Thinking with Generated Images},
  author={Chern, Ethan and Hu, Zhulin and Chern, Steffi and Kou, Siqi and Su, Jiadi and Ma, Yan and Deng, Zhijie and Liu, Pengfei},
  journal={arXiv preprint arXiv:2505.22525},
  year={2025}
}

@article{gu2025thinkmorph,
  title={ThinkMorph: Emergent Properties in Multimodal Interleaved Chain-of-Thought Reasoning},
  author={Gu, Jiawei and Hao, Yunzhuo and Wang, Huichen Will and Li, Linjie and Shieh, Michael Qizhe and Choi, Yejin and Krishna, Ranjay and Cheng, Yu},
  journal={arXiv preprint arXiv:2510.27492},
  year={2025}
}

@misc{cheng2026omnir1unifiedgenerativeparadigm,
      title={Omni-R1: Towards the Unified Generative Paradigm for Multimodal Reasoning}, 
      author={Dongjie Cheng and Yongqi Li and Zhixin Ma and Hongru Cai and Yupeng Hu and Wenjie Wang and Liqiang Nie and Wenjie Li},
      year={2026},
      eprint={2601.09536},
      archivePrefix={arXiv},
      primaryClass={cs.AI},
      url={https://arxiv.org/abs/2601.09536}, 
}

@article{yang2025machine,
  title={Machine Mental Imagery: Empower Multimodal Reasoning with Latent Visual Tokens}, 
  author={Zeyuan Yang and Xueyang Yu and Delin Chen and Maohao Shen and Chuang Gan},
  journal={arXiv preprint arXiv:2506.17218},
  year={2025},
  eprint={2506.17218},
  archivePrefix={arXiv},
  primaryClass={cs.CV},
  url={https://arxiv.org/abs/2506.17218}, 
}

@misc{li2025lvr,
      title={Latent Visual Reasoning}, 
      author={Bangzheng Li and Ximeng Sun and Jiang Liu and Ze Wang and Jialian Wu and Xiaodong Yu and Hao Chen and Emad Barsoum and Muhao Chen and Zicheng Liu},
      year={2025},
      journal={arXiv preprint arXiv:2509.24251}
}

@inproceedings{wang2025monetreasoninglatentvisual,
      title={Monet: Reasoning in Latent Visual Space Beyond Images and Language}, 
      author={Qixun Wang and Yang Shi and Yifei Wang and Yuanxing Zhang and Pengfei Wan and Kun Gai and Xianghua Ying and Yisen Wang},
      year={2026},
      booktitle={CVPR}
}

@article{jiang2026geditbench,
  title={GEditBench v2: A Human-Aligned Benchmark for General Image Editing},
  author={Jiang, Zhangqi and Sun, Zheng and Zeng, Xianfang and Yang, Yufeng and Zhang, Xuanyang and Wu, Yongliang and Cheng, Wei and Yu, Gang and Yang, Xu and Wen, Bihan},
  journal={arXiv preprint arXiv:2603.28547},
  year={2026}
}

@article{li2026thinking,
  title={Thinking in Frames: How Visual Context and Test-Time Scaling Empower Video Reasoning},
  author={Li, Chengzu and Wang, Zanyi and Li, Jiaang and Xu, Yi and Zhou, Han and Zhang, Huanyu and An, Ruichuan and Jiang, Dengyang and An, Zhaochong and Vuli{\'c}, Ivan and others},
  journal={arXiv preprint arXiv:2601.21037},
  year={2026}
}

@inproceedings{yang2019my,
  title={Where is my mirror?},
  author={Yang, Xin and Mei, Haiyang and Xu, Ke and Wei, Xiaopeng and Yin, Baocai and Lau, Rynson WH},
  booktitle={Proceedings of the IEEE/CVF international conference on computer vision},
  pages={8809--8818},
  year={2019}
}

@misc{ikenna113_circuitvqadesc2,
  title        = {circuitvqadesc2 Dataset},
  author       = {{IKENNA113}},
  year         = {2024},
  howpublished = {\url{https://huggingface.co/datasets/IKENNA113/circuitvqadesc2}},
  note         = {Accessed: 2026-05},
}

@article{corbiere2025drivingvqa,
  title={DRIVINGVQA: Analyzing Visual Chain-of-Thought Reasoning of Vision Language Models in Real-World Scenarios with Driving Theory Tests},
  author={Corbi{\`e}re, Charles and Roburin, Simon and Montariol, Syrielle and Bosselut, Antoine and Alahi, Alexandre},
  journal={arXiv preprint arXiv:2501.04671},
  year={2025}
}

@inproceedings{zhang2019raven, 
    title={RAVEN: A Dataset for Relational and Analogical Visual rEasoNing}, 
    author={Zhang, Chi and Gao, Feng and Jia, Baoxiong and Zhu, Yixin and Zhu, Song-Chun}, 
    booktitle={Proceedings of the IEEE Conference on Computer Vision and Pattern Recognition (CVPR)}, 
    year={2019}
}

@misc{shi2025mathcanvasintrinsicvisualchainofthought,
      title={MathCanvas: Intrinsic Visual Chain-of-Thought for Multimodal Mathematical Reasoning}, 
      author={Weikang Shi and Aldrich Yu and Rongyao Fang and Houxing Ren and Ke Wang and Aojun Zhou and Changyao Tian and Xinyu Fu and Yuxuan Hu and Zimu Lu and Linjiang Huang and Si Liu and Rui Liu and Hongsheng Li},
      year={2025},
      eprint={2510.14958},
      archivePrefix={arXiv},
      primaryClass={cs.CV},
      url={https://arxiv.org/abs/2510.14958}, 
}

@misc{vissim_tangram_puzzle,
  title        = {tangram\_puzzle Dataset},
  author       = {{VisSim}},
  year         = {2025},
  howpublished = {\url{https://huggingface.co/datasets/VisSim/tangram_puzzle}},
  note         = {Accessed: 2026-05},
}

@inproceedings{zou2022spot,
  title={Spot-the-difference self-supervised pre-training for anomaly detection and segmentation},
  author={Zou, Yang and Jeong, Jongheon and Pemula, Latha and Zhang, Dongqing and Dabeer, Onkar},
  booktitle={European Conference on Computer Vision},
  pages={392--408},
  year={2022},
  organization={Springer}
}

@misc{kingsoft_qzhou_flowchart_qa,
  title        = {QZhou-Flowchart-QA Dataset},
  author       = {{Kingsoft-LLM}},
  year         = {2024},
  howpublished = {\url{https://huggingface.co/datasets/Kingsoft-LLM/QZhou-Flowchart-QA}},
  note         = {Accessed: 2026-05},
}

@inproceedings{wu2025vsp,
  title={Vsp: Diagnosing the dual challenges of perception and reasoning in spatial planning tasks for mllms},
  author={Wu, Qiucheng and Zhao, Handong and Saxon, Michael and Bui, Trung and Wang, William Yang and Zhang, Yang and Chang, Shiyu},
  booktitle={Proceedings of the IEEE/CVF International Conference on Computer Vision},
  pages={2270--2280},
  year={2025}
}

@article{ma2026thinking,
  title={Thinking with Blueprints: Assisting Vision-Language Models in Spatial Reasoning via Structured Object Representation},
  author={Ma, Weijian and Sun, Shizhao and Yu, Tianyu and Wang, Ruiyu and Chua, Tat-Seng and Bian, Jiang},
  journal={arXiv preprint arXiv:2601.01984},
  year={2026}
}

@article{yu2026avic,
  title={When and How Much to Imagine: Adaptive Test-Time Scaling with World Models for Visual Spatial Reasoning},
  author={Yu, Shoubin and Zhang, Yue and Wang, Zun and Yoon, Jaehong and Yao, Huaxiu and Ding, Mingyu and Bansal, Mohit},
  journal={arXiv preprint arXiv:2602.08236},
  year={2026}
}

@article{zhang2026adaptmmbench,
  title={AdaptMMBench: Benchmarking Adaptive Multimodal Reasoning for Mode Selection and Reasoning Process},
  author={Zhang, Xintong and Zhang, Xiaowen and Wu, Jingrong and others},
  journal={arXiv preprint arXiv:2602.02676},
  year={2026}
}

@article{li2026reliable,
  title={Reliable Thinking with Images},
  author={Li, Haobin and Yang, Yutong and Lin, Yijie and Dai, Xiang and Yang, Mouxing and Peng, Xi},
  journal={arXiv preprint arXiv:2602.12916},
  year={2026}
}

@article{zhou2026prm,
  title={What, Whether and How? Unveiling Process Reward Models for Thinking with Images Reasoning},
  author={Zhou, Yujin and Wen, Pengcheng and Chen, Jiale and others},
  journal={arXiv preprint arXiv:2602.08346},
  year={2026}
}

@article{li2026viebench,
  title={Beyond Accuracy: Evaluating Grounded Visual Evidence in Thinking with Images},
  author={Li, Xuchen and Li, Xuzhao and Pi, Renjie and Hu, Shiyu and Zhao, Jian and Gao, Jiahui},
  journal={arXiv preprint arXiv:2601.11633},
  year={2026}
}

@article{zhou2026genvcot,
  title={Gen-VCoT: Generative Visual Chain-of-Thought Reasoning via Diffusion-Based RGB Intermediate Representations},
  author={Zhou, Zhiqiang and Dai, Junliang and Ling, Xu},
  journal={arXiv preprint arXiv:2606.16783},
  year={2026}
}

@article{li2026visualopsd,
  title={Visual-OPSD: Cross-Modal On-Policy Self-Distillation for Efficient Unified Multimodal Reasoning},
  author={Li, Pengyu and Gao, Zhitao and Zhang, Lingling and Huang, Muye and Li, Yuanming and Xu, Fangzhi and Liu, Jun},
  journal={arXiv preprint arXiv:2606.18974},
  year={2026}
}

@article{wang2026illusion,
  title={The Illusion of Visual Tool-Use: A Causal Audit of Thinking with Images},
  author={Wang, Zhiheng and Peng, Bo and Wei, Lai and Lu, Chaochao},
  journal={arXiv preprint arXiv:2608.06270},
  year={2026}
}

@article{mao2026toolvision,
  title={ToolVision: Learning When and How to Use Visual Tools with Capability-Aligned Supervision},
  author={Mao, Delin and Sun, Chenghao and Song, Jingwei and Chen, Chishui and Zhang, Linfeng},
  journal={arXiv preprint arXiv:2608.08907},
  year={2026}
}

@article{xiang2026openvistool,
  title={OpenVisTool: An Open Recipe for Synthesizing Instructive Visual Tool-Use Trajectories},
  author={Xiang, Changhao and Zhang, Shilin and Ma, Zheng and others},
  journal={arXiv preprint arXiv:2608.08557},
  year={2026}
}

@article{shao2026textcall,
  title={Thinking With Tools, Not With Pixels: Tool Calls as Text Scaffolds for Visual Reasoning},
  author={Shao, Jiahao and Yang, Yuanbo and Liao, Yiyi and Shen, Yujun and Yang, Ceyuan and Xu, Yinghao},
  journal={arXiv preprint arXiv:2608.09682},
  year={2026}
}

@article{zhou2026visualplanning,
  title={Probing Visual Planning in Image Editing Models},
  author={Zhou, Zhimu and Zhao, Yanpeng and Liao, Qiuyu and Zhao, Bo and Ma, Xiaojian},
  journal={arXiv preprint arXiv:2604.22868},
  year={2026}
}

@article{sugiyama2026wisrd,
  title={Image-Space Rule Discovery},
  author={Sugiyama, Misora and Oyama, Toya and Kataoka, Hirokatsu},
  journal={arXiv preprint arXiv:2608.00490},
  year={2026}
}

@article{janjua2026perceptionprograms,
  title={Don't Show Pixels, Show Cues: Unlocking Visual Tool Reasoning in Language Models via Perception Programs},
  author={Janjua, Muhammad Kamran and Silva, Hugo and Niu, Di and Rashidi, Bahador},
  journal={arXiv preprint arXiv:2604.12896},
  year={2026}
}

@article{zeller2026mentisoculi,
  title={MentisOculi: Revealing the Limits of Reasoning with Mental Imagery},
  author={Zeller, Jana and Wiedemer, Thadd{\"a}us and Li, Fanfei and others},
  journal={arXiv preprint arXiv:2602.02465},
  year={2026}
}

@inproceedings{andreas2016neural,
  title={Neural Module Networks},
  author={Andreas, Jacob and Rohrbach, Marcus and Darrell, Trevor and Klein, Dan},
  booktitle={Proceedings of the IEEE Conference on Computer Vision and Pattern Recognition},
  pages={39--48},
  year={2016}
}

@inproceedings{johnson2017clevr,
  title={{CLEVR}: A Diagnostic Dataset for Compositional Language and Elementary Visual Reasoning},
  author={Johnson, Justin and Hariharan, Bharath and van der Maaten, Laurens and Fei-Fei, Li and Zitnick, C. Lawrence and Girshick, Ross},
  booktitle={Proceedings of the IEEE Conference on Computer Vision and Pattern Recognition},
  pages={2901--2910},
  year={2017}
}

@inproceedings{johnson2017inferring,
  title={Inferring and Executing Programs for Visual Reasoning},
  author={Johnson, Justin and Hariharan, Bharath and van der Maaten, Laurens and Hoffman, Judy and Fei-Fei, Li and Zitnick, C. Lawrence and Girshick, Ross},
  booktitle={Proceedings of the IEEE International Conference on Computer Vision},
  pages={2989--2998},
  year={2017}
}

@inproceedings{weber2017imagination,
  title={Imagination-Augmented Agents for Deep Reinforcement Learning},
  author={Weber, Th{\'e}ophane and Racani{\`e}re, S{\'e}bastien and Reichert, David P. and Buesing, Lars and Guez, Arthur and Jimenez Rezende, Danilo and Badia, Adria Puigdomenech and Vinyals, Oriol and Heess, Nicolas and Li, Yujia and Pascanu, Razvan and Battaglia, Peter and Hassabis, Demis and Silver, David and Wierstra, Daan},
  booktitle={Advances in Neural Information Processing Systems},
  year={2017}
}

@article{ebert2018visualforesight,
  title={Visual Foresight: Model-Based Deep Reinforcement Learning for Vision-Based Robotic Control},
  author={Ebert, Frederik and Finn, Chelsea and Lee, Alex X. and Levine, Sergey},
  journal={arXiv preprint arXiv:1812.00568},
  year={2018}
}

@article{ha2018worldmodels,
  title={World Models},
  author={Ha, David and Schmidhuber, J{\"u}rgen},
  journal={arXiv preprint arXiv:1803.10122},
  year={2018}
}

@inproceedings{isola2017image,
  title={Image-to-Image Translation with Conditional Adversarial Networks},
  author={Isola, Phillip and Zhu, Jun-Yan and Zhou, Tinghui and Efros, Alexei A.},
  booktitle={Proceedings of the IEEE Conference on Computer Vision and Pattern Recognition},
  pages={1125--1134},
  year={2017}
}

@inproceedings{zhu2017unpaired,
  title={Unpaired Image-to-Image Translation using Cycle-Consistent Adversarial Networks},
  author={Zhu, Jun-Yan and Park, Taesung and Isola, Phillip and Efros, Alexei A.},
  booktitle={Proceedings of the IEEE International Conference on Computer Vision},
  pages={2223--2232},
  year={2017}
}

@inproceedings{zhu2016generative,
  title={Generative Visual Manipulation on the Natural Image Manifold},
  author={Zhu, Jun-Yan and Kr{\"a}henb{\"u}hl, Philipp and Shechtman, Eli and Efros, Alexei A.},
  booktitle={Proceedings of the European Conference on Computer Vision},
  pages={597--613},
  year={2016}
}

@inproceedings{meng2022sdedit,
  title={{SDEdit}: Guided Image Synthesis and Editing with Stochastic Differential Equations},
  author={Meng, Chenlin and He, Yutong and Song, Yang and Song, Jiaming and Wu, Jiajun and Zhu, Jun-Yan and Ermon, Stefano},
  booktitle={International Conference on Learning Representations},
  year={2022}
}

@inproceedings{hertz2023prompt,
  title={Prompt-to-Prompt Image Editing with Cross-Attention Control},
  author={Hertz, Amir and Mokady, Ron and Tenenbaum, Jay and Aberman, Kfir and Pritch, Yael and Cohen-Or, Daniel},
  booktitle={International Conference on Learning Representations},
  year={2023}
}

@inproceedings{brooks2023instructpix2pix,
  title={InstructPix2Pix: Learning to Follow Image Editing Instructions},
  author={Brooks, Tim and Holynski, Aleksander and Efros, Alexei A.},
  booktitle={Proceedings of the IEEE/CVF Conference on Computer Vision and Pattern Recognition},
  pages={18392--18402},
  year={2023}
}

@inproceedings{zhang2024magicbrush,
  title={MagicBrush: A Manually Annotated Dataset for Instruction-Guided Image Editing},
  author={Zhang, Kai and Mo, Lingbo and Chen, Wenhu and Sun, Huan and Su, Yu},
  booktitle={Advances in Neural Information Processing Systems},
  year={2023}
}

@inproceedings{chang2025oneigbench,
  title={{OneIG-Bench}: Omni-Dimensional Nuanced Evaluation for Image Generation},
  author={Chang, Jingjing and Fang, Yixiao and Xing, Peng and Wu, Shuhan and Cheng, Wei and Wang, Rui and Zeng, Xianfang and Yu, Gang and Chen, Hai-Bao},
  booktitle={Advances in Neural Information Processing Systems},
  year={2025}
}

@inproceedings{yang2025omnisvg,
  title={{OmniSVG}: A Unified Scalable Vector Graphics Generation Model},
  author={Yang, Yiying and Cheng, Wei and Chen, Sijin and Zeng, Xianfang and Yin, Fukun and Zhang, Jiaxu and Wang, Liao and Yu, Gang and Ma, Xingjun and Jiang, Yu-Gang},
  booktitle={Advances in Neural Information Processing Systems},
  year={2025}
}

@article{chen2025regione,
  title={{RegionE}: Adaptive Region-Aware Generation for Efficient Image Editing},
  author={Chen, Pengtao and Zeng, Xianfang and Zhao, Maosen and Shen, Mingzhu and Ye, Peng and Xiang, Bangyin and Wang, Zhibo and Cheng, Wei and Yu, Gang and Chen, Tao},
  journal={arXiv preprint arXiv:2510.25590},
  year={2025}
}

@article{yang2026realrestorer,
  title={{RealRestorer}: Towards Generalizable Real-World Image Restoration with Large-Scale Image Editing Models},
  author={Yang, Yufeng and Zeng, Xianfang and Jiang, Zhangqi and Yin, Fukun and Liu, Jianzhuang and Cheng, Wei and Lan, Jinghong and Liu, Shiyu and Peng, Yuqi and Yu, Gang and Chen, Shifeng},
  journal={arXiv preprint arXiv:2603.25502},
  year={2026}
}

@article{fu2025imontage,
  title={{iMontage}: Unified, Versatile, Highly Dynamic Many-to-Many Image Generation},
  author={Fu, Zhoujie and Zeng, Xianfang and Lan, Jinghong and Liao, Xinyao and Chen, Cheng and Chen, Junyi and Wei, Jiacheng and Cheng, Wei and Liu, Shiyu and Chen, Yunuo and Yu, Gang and Lin, Guosheng},
  journal={arXiv preprint arXiv:2511.20635},
  year={2025}
}

@article{xu2025withanyone,
  title={{WithAnyone}: Towards Controllable and ID Consistent Image Generation},
  author={Xu, Hengyuan and Cheng, Wei and Xing, Peng and Fang, Yixiao and Wu, Shuhan and Wang, Rui and Zeng, Xianfang and Jiang, Daxin and Yu, Gang and Ma, Xingjun and Jiang, Yu-Gang},
  journal={arXiv preprint arXiv:2510.14975},
  year={2025}
}

@inproceedings{zhuang2026vistorybench,
  title={{ViStoryBench}: Comprehensive Benchmark Suite for Story Visualization},
  author={Zhuang, Cailin and Huang, Ailin and Hu, Yaoqi and Wu, Jingwei and Cheng, Wei and Liao, Jiaqi and Wang, Hongyuan and Liao, Xinyao and Cai, Weiwei and Xu, Hengyuan and Zhang, Xuanyang and Zeng, Xianfang and Huang, Zhewei and Yu, Gang and Zhang, Chi},
  booktitle={Proceedings of the IEEE/CVF Conference on Computer Vision and Pattern Recognition},
  year={2026}
}
